\documentclass[journal]{IEEEtran}
\pdfoutput=1
\usepackage[ruled,vlined,linesnumbered,commentsnumbered]{algorithm2e}
\usepackage{booktabs}
\usepackage{multirow}
\usepackage{url}
\usepackage{amsmath}
\usepackage{amssymb}
\usepackage{mathrsfs}
\usepackage{graphicx}
\usepackage{subfigure}
\usepackage[table]{xcolor}

\usepackage{float}
\usepackage{array}
\usepackage{multirow}
\usepackage{flushend}

\usepackage{rotating,tabularx,longtable}

\makeatletter
\newcommand{\removelatexerror}{\let\@latex@error\@gobble}
\makeatother

\begin{document}

\title{Reinforcement Learning in Healthcare: A Survey}
\author{Chao~Yu, Jiming Liu,~\IEEEmembership{Fellow,~IEEE}, and Shamim Nemati
%\author{(to be confirmed)
\IEEEcompsocitemizethanks{\IEEEcompsocthanksitem Chao~Yu is with the School of Data and Computer Science, Sun Yat-sen University, Guangzhou, China. (Email: yuchao3@mail.sysu.edu.cn).
Jiming Liu is with the Computer Science Department, Hong Kong Baptist University, Kowloon Tong, Hong Kong. (Email: jiming@Comp.HKBU.Edu.HK).
Shamim Nemati is with the Department of Biomedical Informatics, UC San Diego, La Jolla, CA, USA. (Email: snemati@health.ucsd.edu).
\protect\\
}% <-this % stops a space
\thanks{}}

\maketitle
\thispagestyle{empty}

\begin{abstract}
As a subfield of machine learning, \emph{reinforcement learning} (RL) aims at empowering one's capabilities in behavioural decision making by using interaction experience with the world and an evaluative feedback. Unlike traditional supervised learning methods that usually rely on one-shot, exhaustive and supervised reward signals, RL tackles with sequential decision making problems with sampled, evaluative and delayed feedback simultaneously. Such distinctive features make RL technique a suitable candidate for developing powerful solutions in a variety of healthcare domains, where diagnosing decisions or treatment regimes are usually characterized by a prolonged and sequential procedure. This survey discusses the broad applications of RL techniques in healthcare domains, in order to provide the research community with systematic understanding of theoretical foundations, enabling methods and techniques, existing challenges, and new insights of this emerging paradigm. By first briefly examining theoretical foundations and key techniques in RL research from efficient and representational directions, we then provide an overview of RL applications in healthcare domains ranging from dynamic treatment regimes in chronic diseases and critical care, automated medical diagnosis from both unstructured and structured clinical data, as well as many other control or scheduling domains that have infiltrated many aspects of a healthcare system. Finally, we summarize the challenges and open issues in current research, and point out some potential solutions and directions for future research.
\end{abstract}


\IEEEpeerreviewmaketitle

\section{Introduction}\label{sec:introduction}

Driven by the increasing availability of massive multimodality data, and developed computational models and algorithms, the role of AI techniques in healthcare has grown rapidly in the past decade \cite{patel2009coming,dilsizian2014artificial,jiang2017artificial,he2019practical}. This emerging trend has promoted increasing interests in the proposal of advanced data analytical methods and machine learning approaches in a variety of healthcare applications \cite{johnson2016machine,ravi2017deep,ching2018opportunities,luo2016big,esteva2019guide}. As as a subfield in machine learning, \emph{reinforcement learning} (RL) has achieved tremendous theoretical and technical achievements in generalization, representation and efficiency in recent years, leading to its increasing applicability to real-life problems in playing games, robotics control, financial and business management, autonomous driving, natural language processing, computer vision, biological data analysis, and art creation, just to name a few \cite{mnih2015human,littman2015reinforcement,li2018deep,mahmud2018applications,sutton2018reinforcement}.

In RL problems, an agent chooses an action at each time step based on its current state, and receives an evaluative feedback and the new state from the environment. The goal of the agent is to learn an optimal policy (i.e., a mapping from the states to the actions) that maximizes the accumulated reward it receives over time. Therefore, agents in RL do not receive direct instructions regarding which action they should take, instead they must learn which actions are the best through trial-and-error interactions with the environment. This adaptive closed-loop feature renders RL distinct from traditional supervised learning methods for regression or classification, in which a list of correct labels must be provided, or from unsupervised learning approaches to dimensionality reduction or density estimation, which aim at finding hidden structures in a collection of example data \cite{littman2015reinforcement}. Moreover, in comparison with other traditional control-based methods, RL does not require a well-represented mathematical model of the environment, but develops a control policy directly from experience to predict states and rewards during a learning procedure. Since the design of RL is letting an agent controller interact with the system, unknown and time-varying dynamics as well as changing performance requirements can be naturally accounted for by the controller \cite{bucsoniu2018reinforcement}. Lastly, RL is uniquely suited to systems with inherent time delays, in which decisions are performed without immediate knowledge of effectiveness, but evaluated by a long-term future reward.

The above features naturally make RL an attractive solution to constructing efficient policies in various healthcare domains, where the decision making process is usually characterized by a prolonged period or sequential procedure \cite{gottesman2019guidelines}. Typically, a medical or clinical treatment regime is composed of a sequence of decision to determine the course of decisions such as treatment type, drug dosage, or re-examination timing at a time point according to the current health status and prior treatment history of an individual patient, with a goal of promoting the patient's long-term benefits. Unlike the common procedure in traditional randomized controlled trials that derive treatment regimes from the average population response, RL can be tailored for achieving precise treatment for individual patients who may possess high heterogeneity in response to the treatment due to variety in disease severity, personal characteristics and drug sensitivity. Moreover, RL is able to find optimal policies using only previous experiences, without requiring any prior knowledge about the mathematical model of the biological systems. This makes RL more appealing than many existing control-based approaches in healthcare domains since it could be usually difficult or even impossible to build an accurate model for the complex human body system and the responses to administered treatments, due to nonlinear, varying and delayed interaction between treatments and human bodies.

Thus far, a plethora of theoretical or experimental studies have applied RL techniques and models in a variety of heathcare domains, achieving performance exceeding that of alternative techniques in many cases. This survey aims at providing an overview of such successful RL applications, covering adaptive treatment regimes in chronic diseases and critical care, automated clinical diagnosis, as well as many other healthcare domains such as clinical resource allocation/scheduling and optimal process control. We also discuss the challenges, open issues and future directions of research necessary to advance further successful applications of RL in healthcare. By this, we hope this survey can provide the research community with systematic understanding of foundations, enabling methods and techniques, challenges, and new insights of this emerging paradigm. Section \ref{sec:RL} provides a structured summarization of the theoretical foundations and key techniques in RL research from two main directions: \emph{efficient} directions that mainly aim at improving learning efficiency by making best use of past experience or knowledge, and \emph{representational} directions that focus on constructive or relational representation problems in RL. Then, Section \ref{sec:RL_Healthcare} gives a global picture of application domains of RL in healthcare, each of which is discussed in more detail in the following sections. Section \ref{sec:RL_DTR} discusses dynamic treatment regimes in both chronic disease and critical care, and Section \ref{sec:diagnosis} describes automated medical diagnosis using either structured or unstructured medical data. In addition, \ref{sec:other_domains} talks about other more broad application domains including health resources allocation and scheduling, optimal process control, drug discovery and development, as well as health management. Section \ref{sec:challenges} describes several challenges and open issues in current research. Finally, Section \ref{sec:future_perspectives} discusses potential directions that are necessary in the future research. For convenience, Tables \ref{tbl:RL_abbreviation} and \ref{tbl:health_abbreviation} summarize the main acronyms in RL and healthcare domains, respectively.

\begin{table}[!tb]
\centering
\caption{Summary of Abbreviations in RL}
  \label{tab:Abbreviations_RL}
  \scalebox{0.88}{
    \begin{tabular}{|p{2cm}|p{5.5cm}|}
        \toprule
        \hline
        Acronym& Description \\\hline
        AC &  \emph{Actor-Critic} \\\hline
        A3C & \emph{Asynchronous Advantage Actor Critic} \\\hline
        BRL &  \emph{Batch Reinforcement Learning} \\\hline
        DDPG& \emph{Deep Deterministic Policy Gradient} \\\hline
        DRL & \emph{Deep Reinforcement Learning} \\\hline
        DP& \emph{Dynamic programming}\\\hline
        DQN &  \emph{Deep Q Network }\\\hline
        DDQN &  \emph{Dueling DQN}\\\hline
        DDDQN &  \emph{Double Dueling DQN}\\\hline
        FQI-SVG/ERT & \emph{Fitted Q Iteration with Support Vector Regression/Extremely Randomized Trees} \\\hline
        GAN & \emph{Generative Adversarial Net}  \\\hline
        HRL & \emph{Hierarchical Reinforcement Learning} \\\hline
        IRL & \emph{Inverse Reinforcement Learning} \\\hline
        LSPI& \emph{Least-Squares Policy Iteration} \\\hline
        MDP & \emph{Markov Decision Process} \\\hline
        MC & \emph{Monte Carlo} \\\hline
        NAC & \emph{Natural Actor Critic}  \\\hline
       PAC& \emph{Probably Approximately Correct} \\\hline
        PI & \emph{Policy Iteration} \\\hline
        PS & \emph{Policy Search} \\\hline
        POMDP & \emph{Partially Observed Markov Decision Process}\\\hline
        PORL   & \emph{Partially Observed Reinforcement Learning} \\\hline
        PPO & \emph{Proximal Policy Optimization}  \\\hline
        PRL &  \emph{Preference-based Reinforcement Learning} \\\hline
        RRL &  \emph{Relational Reinforcement Learning} \\\hline
        TD &  \emph{Temporal Difference} \\\hline
        TRL&  \emph{Transfer Reinforcement Learning} \\\hline
        TRPO& \emph{Trust Region Policy Optimization} \\\hline
        VI & \emph{Value Iteration} \\\hline
     \end{tabular}
     }
       \label{tbl:RL_abbreviation}
\end{table}

\begin{table}[!tb]
\centering
\caption{Summary of Abbreviations in Healthcare}
  \label{tab:Abbreviations_healthcare}
  \scalebox{0.88}{
    \begin{tabular}{|p{2cm}|p{6.5cm}|}
        \toprule
        \hline
        Acronym& Description \\\hline
       AMP & \emph{Anemia Management Protocol}  \\\hline
        BCSC& \emph{Breast Cancer Surveillance Consortium} \\\hline
        CATIE &\emph{Clinical Antipsychotic Trials of Intervention Effectiveness}\\\hline
        CBASP&  \emph{Cognitive Behavioral-Analysis System of Psychotherapy}\\\hline
        CIBMTR & \emph{Center for International Blood and Marrow Transplant Research}  \\\hline
        CT & \emph{Computed Tomography} \\\hline
        DTRs &  \emph{Dynamic Treatment Regimes}  \\\hline
        EEG & \emph{Electroencephalograph} \\\hline
        ESAs & \emph{Erythropoiesis-Stimulating Agents} \\\hline
        EPO & \emph{Endogenous Erythropoietin} \\\hline
        FES &\emph{Functional Electrical Stimulation} \\\hline
        HGB & \emph{Hemoglobin}  \\\hline
        ICU & \emph{Intensive Care Unit}  \\\hline
        MAP & \emph{Mean Arterial Pressure} \\\hline
        MDD & \emph{Major Depressive Disorder} \\\hline
        MIMIC& \emph{Multiparameter Intelligent Monitoring in Intensive Care} \\\hline
        MRI &\emph{ Magnetic Resonance Images} \\\hline
        NSCLC & \emph{Non-small Sell Lung Cancer} \\\hline
        ODE & \emph{Ordinary Difference Equations} \\\hline
        PK/PD & \emph{PharmacoKinetic/PharmacoDynamic} \\\hline
        SC & \emph{Symptom Checking} \\\hline
        SMARTs& \emph{Sequential Multiple Assignment Randomized Trials}  \\\hline
        STAR*D  & \emph{Sequenced Treatment Alternatives to Relieve Depression} \\\hline
        STI &  \emph{Structured Treatment Interruption}   \\\hline
        SUD & \emph{Sub-stance Use Disorder} \\\hline
        TREC-CDS &  \emph{Text REtrieval Conference-Clinical Decision Support}  \\\hline
        UI & \emph{Ultrasound Images}  \\\hline
     \end{tabular}
     }
     \label{tbl:health_abbreviation}
\end{table}

\section{Theoretical Foundations and Key Techniques in RL}\label{sec:RL}
\begin{figure*}[!tb]
\centering
\includegraphics[width=0.8\textwidth]{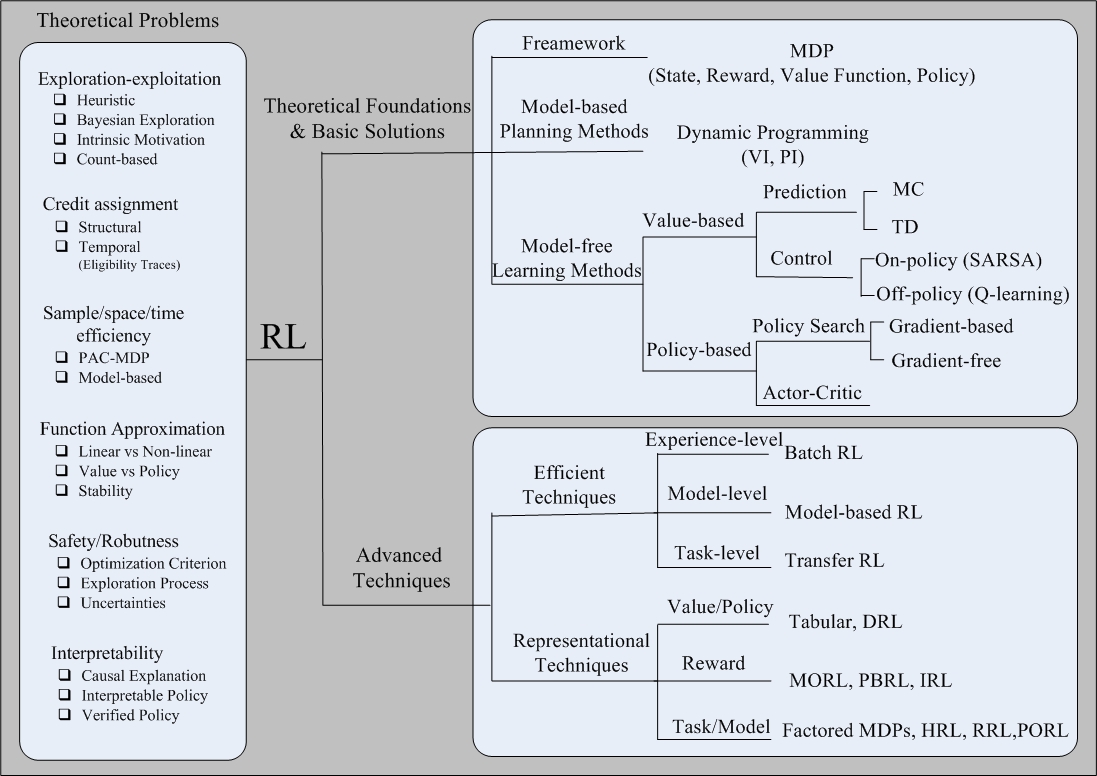}
%\vspace{-0.3cm}
\caption{The summarization of theoretical foundations, basic solutions, challenging issues and advanced techniques in RL.}
\label{fig:RL_overall_framework}
\end{figure*}

This section serves as a brief introduction to the theoretical models, basic solutions and advanced techniques in RL. The goal is to provide a quick overview of what constitutes the main components of RL methods. Some fundamental concepts and major theoretical problems are also clarified. Subsection \ref{subsec:RL_theoretical} first discusses the general decision making framework for RL, and its model-based and model-free solutions. Then, Subsection \ref{subsec:key_techniques} describes some advanced RL techniques from perspectives of facilitating learning efficiency and enriching constructive or relational representation capabilities of RL approaches. Fig.~\ref{fig:RL_overall_framework} provides a diagram outlining the main components and sub-field research topics in RL.

\subsection{Theoretical Foundations of RL}\label{subsec:RL_theoretical}
RL enables an agent to learn effective strategies in sequential decision making problems by trial-and-error interactions with its environment \cite{sutton2018reinforcement}. The \emph{Markov decision process (MDP)}, which has a long history in the research of \emph{theoretic decision making} in stochastic settings, has been used as a general framework to formalize an RL problem. Approaching an MDP can be conducted in various forms depending on what information of the targeted problem can be specified a priori.

\subsubsection{MDP}\label{sec:MDP}
Formally, an MDP can be defined by a 5-tuple $\mathcal{M} = (\mathcal{S}, \mathcal{A}, \mathcal{P}, \mathcal{R},\gamma)$, where $\mathcal{S}$ is a finite \emph{state} space, and $s_t\in \mathcal{S}$ denotes the state of an agent at time $t$; $\mathcal{A}$ is a set of \emph{actions} available to the agent, and $a_t\in \mathcal{A}$ denotes the action that the agent performs at time $t$; $\mathcal{P}(s,a,s^{\prime}): \mathcal{S} \times \mathcal{A} \times \mathcal{S} \rightarrow [0,1]$ is a Markovian \emph{transition function} when the agent transits from state $s$ to state $s^{\prime}$ after taking action $a$; $\mathcal{R}: \mathcal{S} \times \mathcal{A} \rightarrow \Re$ is a \emph{reward function} that returns the immediate reward $\mathcal{R}(s,a)$ to the agent after taking action $a$ in state $s$; and $\gamma \in [0,1]$ is a \emph{discount factor}.

An agent's \emph{policy} $\pi: \mathcal{S} \times \mathcal{A} \rightarrow [0,1]$ is a probability distribution that maps an action $a \in \mathcal{A}$  to a state $s \in \mathcal{S}$. When given an MDP and a policy $\pi$, the expected reward of following this policy when starting in state $s$, $V^{\pi}(s)$, can be defined as follows:

\begin{equation}\label{equ:v}
 V^{\pi}(s) \triangleq E_{\pi} \left[\sum_{t=0}^{\infty}\gamma^{t}\mathcal{R}(s_{t},\pi(s_{t}))|s_{0}=s\right]
\end{equation}

The value function can also be defined recursively using the \emph{Bellman operator $\mathscr{B^\pi}$}:

\begin{equation}\label{equ:bellman_value}
  \mathscr{B^\pi}V^\pi(s) \triangleq \mathcal{R}(s,\pi(s)) + \gamma \sum_{s^{\prime}\in \mathcal{S}}\mathcal{P}(s,a,s^{\prime})V^{\pi}(s^{\prime})
\end{equation}

Since the \emph{Bellman operator $\mathscr{B^\pi}$} is a contraction mapping of value function $V$, there exists a fixed point of value $V^\pi$ such that $\mathscr{B^\pi}V^\pi=V^\pi$ in the limit. The goal of an MDP problem is to compute an \emph{optimal policy} $\pi^{\ast}$ such that $V^{\pi^{\ast}}(s)\geq V^{\pi}(s)$ for every policy $\pi$ and every state $s\in S$. To involve the action information, Q-value is used to represent the optimal value of each state-action pair by Equation \ref{equ:Q_value}.

\begin{equation}\label{equ:Q_value}
 Q^{\ast}(s,a) = \mathcal{R}(s,a) + \gamma \sum_{s^{\prime}\in \mathcal{S}}\mathcal{P}(s,a,s^{\prime})\max_{a^{\prime}\in \mathcal{A}} Q(s^{\prime},a^{\prime})
\end{equation}

\subsubsection{Basic Solutions and Challenging Issues}\label{sec:basic_solutions}

Many solution techniques are available to compute an optimal policy for a given MDP. Broadly, these techniques can be categorized as \emph{model-based} or \emph{model-free} methods, based on whether a complete knowledge of the MDP model can be specified a priori. Model-based methods, also referred to as \emph{planning} methods, require a complete description of the model in terms of the transition and reward functions, while model-free methods, also referred to as \emph{learning} methods, learn an optimal policy simply based on received observations and rewards.

\emph{Dynamic programming} (DP) \cite{bellman2013dynamic} is a collection of \emph{model-based} techniques to compute an optimal policy given a complete description of an MDP model. DP includes two main different approaches: \emph{Value Iteration} (VI) and \emph{Policy Iteration} (PI). VI specifies the optimal policy in terms of value function $Q^{\ast}(s,a)$ by iterating the \emph{Bellman updating} as follows:

\begin{equation}\label{equ:bellman_transformed}
 Q_{t+1}(s,a) =  \mathcal{R}(s,a) + \gamma \sum_{s^{\prime}\in \mathcal{S}}\mathcal{P}(s,a,s^{\prime})\max_{a^{\prime}\in \mathcal{A}} Q_{t}(s^{\prime},a^{\prime})
\end{equation}

For each iteration, the value function of every state $s$ is updated one step further into the future based on the current estimate. The concept of updating an estimate based on the basis of other estimates is often referred to as \emph{bootstrapping}. The value function is updated until the difference between two iterations, $Q_t$ and $Q_{t+1}$, is less than a small threshold. The optimal policy is then derived using $\pi^{\ast}(s)= \arg \max _{a\in \mathcal{A}} Q^{\ast}$. Unlike VI, PI learns the policy directly. It starts with an initial random policy $\pi$, and iteratively updates the policy by first computing the associated value function $Q^{\pi}$ (\emph{policy evaluation} or \emph{prediction}) and then improving the policy using $\pi(s)= \arg \max _{a\in \mathcal{A}} Q(s,a)$ (\emph{policy improvement}  or \emph{control}).

Despite being mathematically sound, DP methods require a complete and accurate description of the environment model, which is unrealistic in most applications. When a model of the problem is not available, the problem can then be solved by using \emph{direct RL} methods, in which an agent learns its optimal policy while interacting with the environment. \emph{Monte Carlo} (MC) methods and \emph{Temporal difference} (TD) methods are two main such methods, with the difference of using episode-by-episode update in MC or step-by-step update in TD. Denote $\mathcal{R}_t^{(n)}=\mathcal{R}_{t+1}+\gamma\mathcal{R}_{t+2}+...+\gamma^{n-1}\mathcal{R}_{t+n}+\gamma^nV_t(s_{t+n})$ $\emph{n}$-step return at time t, then the general $\emph{n}-$step update rule in TD methods is defined by $\Delta V_t(s_t)=\alpha[\mathcal{R}_t^{(n)}-V_t(s_t)]$, in which $\alpha\in(0,1]$ is an appropriate learning rate controlling the contribution of the new experience to the current estimate. MC methods then can be considered as an extreme case of TD methods when the update is conducted after the whole episode of steps. In spite of having higher complexity in analyzing the efficiency and speed of convergence, TD methods usually require less memory for estimates and less computation, thus are easier to implement.

If the value function of a policy $\pi$ is estimated by using samples that are generated by strictly following this policy, the RL algorithm is called \emph{on-policy}, while \emph{off-policy} algorithms can learn the value of a policy that is different from the one being followed. One of the most important and widely used RL approach is Q-learning \cite{watkins1992q}, which is an \emph{off-policy} TD algorithm. Its one-step updating rule is given by Equation \ref{equ:Q1},

\begin{equation}\label{equ:Q1}
 Q_{t+1}(s,a) = Q_{t}(s,a) + \alpha_{t}[R(s,a) + \gamma \max_{a^{\prime}} Q_t(s^{\prime},a^{\prime}) - Q_t(s,a)]
\end{equation}
where  $\alpha\in(0,1]$ is an appropriate learning rate which controls the contribution of the new experience to the current estimate.

Likewise, the SARSA algorithm \cite{rummery1994line} is an representation for \emph{on-policy} TD approaches given by Equation \ref{equ:SARSA}:
\begin{equation}\label{equ:SARSA}
 Q_{t+1}(s,a) = Q_{t}(s,a) + \alpha_{t}[R(s,a) + \gamma Q_t(s^{\prime},\pi(s^{\prime})) - Q_t(s,a)]
\end{equation}

The idea is that each experienced sample brings the current estimate $Q(s,a)$ closer to the optimal value $Q^{\ast}(s,a)$. Q-learning starts with an initial estimate for each state-action pair. When an action $a$ is taken in state $s$, resulting in the next state $s^{\prime}$, the corresponding Q-value $Q(s,a)$ is updated with a combination of its current value and the TD error ( $R(s,a) + \gamma \max_{a^{\prime}} Q_t(s^{\prime},a^{\prime}) - Q_t(s,a)$ for Q-learning, or $R(s,a) + \gamma Q_t(s^{\prime},\pi(s^{\prime})) - Q_t(s,a)$ for SARSA). The TD error is the difference between the current estimate $Q(s,a)$ and the expected discounted return based on the experienced sample. The $Q$ value of each state-action pair is stored in a table for a discrete state-action space. It has been proved that this tabular Q-learning converges to the optimal $Q^{\ast}(s,a)$ w.p.1 when all state-action pairs are visited infinitely often and an appropriate exploration strategy and learning rate are chosen~\cite{watkins1992q}.

Besides the above \emph{value-function} based methods that maintain a value function whereby a policy can be derived, \emph{direct policy-search} (PS) algorithms \cite{kober2009policy} try to estimate the policy directly without representing a value function explicitly, whereas the \emph{actor-critic} (AC) methods \cite{peters2008natural} keep separate, explicit representations of both value functions and policies. In AC methods, the \emph{actor} is the policy to select actions, and the \emph{critic} is an estimated value function to criticize the actions chosen by the actor. After each action execution, the critic evaluates the performance of action using the TD error. The advantages of AC methods include that they are more appealing in dealing with large scale or even continuous actions and learning stochastic policies, and more easier in integrating domain specific constraints on policies.

In order to learn optimal policies, an RL agent should make a balance between exploiting the knowledge obtained so far by acting optimally, and exploring the unknown space in order to find new efficient actions. Such an \emph{exploration-exploitation trade-off} dilemma is one of the most fundamental theoretical issues in RL, since an effective \emph{exploration strategy} enables the agent to make an elegant balance between these two processes by choosing explorative actions only when this behavior can potentially bring a higher expected return. A large amount of effort has been devoted to this issue in the traditional RL community, proposing a wealth of exploration strategies including simple heuristics such as $\varepsilon$-greedy and Boltzmann exploration, Bayesian learning \cite{vlassis2012bayesian,ghavamzadeh2015bayesian}, count-based methods with \emph{Probably Approximately Correct} (PAC) guarantees \cite{strehl2009reinforcement,brafman2002r}, as well as more expressive methods of intrinsic motivation such as novelty, curiosity and surprise \cite{barto2013intrinsic}. For example, the $\varepsilon$-greedy strategy selects the greedy action, $\arg \max_{a} Q_t(s,a)$, with a high probability, and, occasionally, with a small probability selects an action uniformly at random. This ensures that all actions and their effects are experienced. The $\varepsilon$-greedy exploration policy can be given by Equation \ref{equ:exploration}.

\begin{equation} \label{equ:exploration}
\pi(a^{\prime}) = \left\{
               \begin{array}{ll}
          \displaystyle 1-\varepsilon & \hbox{if  $a = \arg \max _{a^{\prime}} Q(s,a)$,} \\
                 \varepsilon & \hbox{  otherwise.}
               \end{array}
             \right.
\end{equation}
where $\varepsilon \in [0,1]$ is an exploration rate.

Other fundamental issues in RL research include but are not limited to the credit assignment problem \cite{sutton2018reinforcement,busoniu2008comprehensive}, the sampel/space/time complexity \cite{kakade2003sample,li2012sample}, function approximation \cite{busoniu2010reinforcement,van2012reinforcement}, safety \cite{moldovan2012safe,garcia2015comprehensive}, robustness \cite{wiesemann2013robust,xu2010distributionally}, and interpretability \cite{hein2018interpretable,bastani2018verifiable}. A more comprehensive and in-depth review on these issues can be found in \cite{wiering2012reinforcement}, and more recently in \cite{li2018deep,sutton2018reinforcement}.

\subsection{Key Techniques in RL}\label{subsec:key_techniques}
This section discusses some key techniques used in contemporary RL, most of which can be understood in the light of the framework and solutions defined in the section ahead, yet these new techniques emphasize more sophisticated use of samples, models of the world and learned knowledge of previous tasks for efficiency purpose, as well as what should be represented and how things should be represented during an RL problem. Note that the classification of these two kinds of techniques are not mutually exclusive, which means that some representation techniques are also used for improving the learning efficiency, and vice versa.
\subsubsection{Efficient Techniques}\label{subsubsec:efficient_techniques}
The purpose of using efficient techniques is to improve the learning performance in terms of, for example, convergence ratio, sample efficient, computation cost or generalization capabilities of an RL method. This improvement can be achieved by using different levels of knowledge: the \emph{Experience-level} techniques focus on utilizing the past experience for more stable and data-efficient learning; the \emph{Model-level} techniques focus on building and planning over a model of the environment in order to improve sample efficiency; while the \emph{Task-level} techniques aim at generalizing the learning experience from past tasks to new relevant ones.

\paragraph{Experience-level}  In traditional pure on-line TD learning methods such as Q-learning and SARSA, an agent immediately conducts a DP-like update of the value functions every step interacting with the environment and then disregards the experienced state transition tuple afterwards. In spite of guaranteed convergence and great success in solving simple toy problems, this kind of \emph{local updates} poses several severe performance problems when applied to more realistic systems with larger and possibly continuous settings. Since each experience tuple is used only for one update and then forgotten immediately, a larger number of samples are required to enable an optimal solution, causing the so called \emph{exploration overhead} problem. Moreover, it has been shown that directly combining function approximation methods with pure on-line TD methods can cause instable or even diverged performance \cite{busoniu2010reinforcement,van2012reinforcement}. These inefficiency and instability problems become even more pronounced in real environments, particularly in healthcare systems, where physical interactions between patients and environments call for more efficient sampling and stable learning methods.

The \emph{Experience-level} techniques focus on how to make the best of the past learning experience for more stable and efficient learning, and are the major driving force behind the proposal of modern \emph{Batch RL (BRL)} \cite{lange2012batch}. In BRL, two basic techniques are used: storing the experience in a buffer and reusing it as if it were new (the idea of \emph{experience replay} for addressing the inefficiency problem), and separating the DP step from the function approximation step by using a supervised learning to fit the function approximator over the sampled experience (the idea of \emph{fitting} for addressing the instability problem). There are several famous BRL approaches in the literature, such as the non-linear approximator cases of \emph{Neural Fitted Q Iteration} (NFQI \cite{riedmiller2005neural}), the Tree-based FQI \cite{ernst2005tree}, and robust linear approximation techniques for policy learning such as \emph{Least-Squares Policy Iteration} (LSPI \cite{lagoudakis2003least}). As will be discovered later, these BRL methods have enjoyed wide and successful applications in clinical decision makings, due to their promise in greatly improving learning speed and approximation accuracy, particularly from limited amounts of clinical data.

\paragraph{Model-level} Unlike \emph{Experience-level} techniques that emphasize the efficient use of experience tuples, the \emph{Model-level} techniques try to build a model of the environment (in terms of the transition and reward functions) and then derive optimal policies from the environment model when it is approximately correct. This kind of \emph{model-based RL} (MRL) approaches is rather different from the model-free RL methods such as TD methods or MC methods that directly estimate value functions without building a model of the environment \cite{hester2012learning}. Using some advanced exploration strategies and planning methods such as DP or \emph{Monte Carlo Tree Search} (MCTS) \cite{browne2012survey}, \emph{MRL} methods are usually able to learn an accurate model quickly and then use this model to plan multi-step actions. Therefore, \emph{MRL} methods normally have better sample efficiency than model-free methods \cite{kakade2003sample}.

%and Knows What It Knows (KWIK) \cite{li2008knows}
\paragraph{Task-level} A higher task-level of efficient approaches focuses on the development of methods to transfer knowledge from a set of source tasks to a target task. \emph{Transfer RL} (TRL) uses the transferred knowledge to significantly improve the learning performance in the target task, e.g., by reducing the samples needed for a nearly optimal performance, or increasing the final convergence level \cite{lazaric2012transfer}. Taylor and Stone \cite{taylor2009transfer} provided a thorough review on TRL approaches by five transfer dimensions: how the source task and target task may differ (e.g., in terms of action, state, reward or transition functions), how to select the source task (e.g., all previously seen tasks, or only one task specified by human or modified automatically), how to define task mappings (e.g., specified by human or learned from experience), what knowledge to transferred (from experience instances to higher level of models or rules), and allowed RL methods (e.g., MRL, PS, or BRL).

\subsubsection{Representational Techniques}\label{subsubsec:representational_techniques}
Unlike traditional machine learning research that simply focuses on feature engineering for function approximation, representational techniques in RL can be in a broader perspective, paying attention to constructive or relational representation problems relevant not only to function approximation for state/action, polices and value functions, but also to more exogenous aspects regarding agents, tasks or models \cite{li2018deep}.

\paragraph{Representation for Value Functions or Policies} Many traditional RL algorithms have been mainly designed for problems with small discrete state and action spaces, which can be explicitly stored in tables. Despite the inherent challenges, applying these RL algorithms to continuous or highly dimensional domains would cause extra difficulties. A major aspect of representational techniques is to represent structures of policies and value functions in a more compact form for an efficient approximation of solutions and thus scaling up to larger domains. Broadly, three categories of approximation methods can be clarified \cite{van2012reinforcement}: \emph{model-approximation} methods that approximate the model and compute the desired policy on this approximated model; \emph{value-approximation} methods that approximate a value function whereby a policy can be inferred, and \emph{policy-approximation} methods that search in policy space directly and update this policy to approximate the optimal policy, or keep separate, explicit representations of both value functions and policies.

The value functions or policies can be parameterized using either \emph{linear} or \emph{non-linear} function approximation presentations. Whereas the \emph{linear function approximation} is better understood, simple to implement and usually has better convergence guarantees, it needs explicit knowledge about domain features, and also prohibits the representation of interactions between features. On the contrary, \emph{non-linear function approximation} methods do not need for good informative features and usually obtain better accuracy and performance in practice, but with less convergence guarantees.

A notable success of RL in addressing real world complex problems is the recent integration of deep neural networks into RL \cite{silver2016mastering,liu2017survey}, fostering a new flourishing research area of \emph{Deep RL} (DRL) \cite{li2018deep}. A key factor in this success is that deep learning can automatically abstract and extract high-level features and semantic interpretation directly from the input data, avoiding complex feature engineering or delicate feature hand-crafting and selection for an individual task \cite{sze2017efficient}.

\paragraph{Representation for Reward Functions}
In a general RL setting, the reward function is represented in the form of an evaluative scalar signal, which encodes a single objective for the learning agent. In spite of its wide applicability, this kind of quantifying reward functions has its limits inevitably. For example, real life problems usually involve two or more objectives at the same time, each with its own associated reward signal. This has motivated the emerging research topic of \emph{ multi-objective RL} (MORL) \cite{liu2015multiobjective}, in which a policy must try to make a trade-off between distinct objectives in order to achieve a Pareto optimal solution. Moreover, it is often difficult or even impossible to obtain feedback signals that can be expressed in numerical rewards in some real-world domains. Instead, qualitative reward signals such as being better or higher may be readily available and thus can be directly used by the learner. \emph{Preference-based RL} (PRL) \cite{wirth2017survey} is a novel research direction combining RL and preference learning \cite{furnkranz2012preference} to equip an RL agent with a capability to learn desired policies from qualitative feedback that is expressed by various ranking functions. Last but not the least, all the existing RL methods are grounded on an available feedback function, either in an explicitly numerical or a qualitative form. However, when such feedback information is not readily available or the reward function is difficult to specify manually, it is then necessary to consider an approach to RL whereby the reward function can be learned from a set of presumably optimal trajectories so that the reward is consistent with the observed behaviors. The problem of deriving a reward function from observed behavior is referred to as \emph{Inverse RL} (IRL) \cite{ng2000algorithms,zhifei2012survey}, which has received an increasingly high interest by researchers in the past few years. Numerous IRL methods have been proposed, including the Maximum Entropy IRL \cite{ziebart2008maximum}, the Apprenticeship Learning \cite{abbeel2004apprenticeship}, nonlinear representations of the reward function using Gaussian processes \cite{levine2011nonlinear}, and Bayesian IRL  \cite{ramachandran2007bayesian}.

\paragraph{Representation for Tasks or Models}
Much recent research on RL has focused on representing the tasks or models in a compact way to facilitate construction of an efficient policy. Factored MDPs \cite{guestrin2003efficient} are one of such approaches to representing large structured MDPs compactly, by using a \emph{dynamic Bayesian network} (DBN) to represent the transition model among states that involve only some set of state variables, and the decomposition of global task reward to individual variables or small clusters of variables. This representation often allows an exponential reduction in the representation size of structured MDPs, but the complexity of exact solution algorithms for such MDPs also grows exponentially in the representation size. A large number of methods has been proposed to employ factored representation of MDP models for improving learning efficiency for either model-based \cite{kearns1999efficient,guestrin2002algorithm} or model-free RL problems \cite{osband2014near}. A more challenging issues is how to learn this compact structure dynamically during on-line learning \cite{strehl2007efficient}.

Besides the factored representation of states, a more general method is to decompose large complex tasks into smaller sets of sub-tasks, which can be solved separatively. \emph{Hierarchical RL} (HRL) \cite{barto2003recent} formalizes hierarchical methods that use abstract states or actions over a hierarchy of subtasks to decompose the original problem, potentially reducing its computational complexity. Hengst \cite{hengst2012hierarchical} discussed the various concepts and approaches in HRL, including algorithms that can automatically learn the hierarchical structure from interactions with the domain. Unlike HRL that focuses on hierarchical decomposition of tasks, \emph{Relational RL} (RRL) \cite{van2012solving} provides a new representational paradigm to RL in worlds explicitly modeled in terms of objects and their relations. Using expressive data structures that represent the objects and relations in an explicit way, RRL aims at generalizing or facilitating learning over worlds with the same or different objects and relations. The main representation methods and techniques in RRL have been surveyed in detail in \cite{van2012solving}.

Last but not the least, \emph{Partially Observable MDP} (POMDP) is widely adopted to represent models when the states are not fully observable, or the observations are noisy. Learning in POMDP, denoted as \emph{Partially Observable RL} (PORL), can be rather difficult due to extra uncertainties caused by the mappings from observations to hidden states \cite{jaakkola1995reinforcement}. Since environmental states in many real life applications, notably in healthcare systems, are only partially observable, PORL then becomes a suitable technique to derive a meaningful policy in such realistic environments.

\section{Applications of RL in Healthcare}\label{sec:RL_Healthcare}
On account of its unique features against traditional machine learning, statistic learning and control-based methods, RL-related models and approaches have been widely applied in healthcare domains since decades ago. The early days of focus has been devoted to the application of DP methods in various pharmacotherapeutic decision making problems using \emph{pharmacokinetic}/\emph{pharmacodynamic} (PK/PD) models \cite{jelliffe1970computer,bellman1983mathematical}. Hu \emph{et al.}, \cite{hu1994comparison} used POMDP to model drug infusion problem for the administration of anesthesia, and proposed efficient heuristics to compute suboptimal though useful treatment strategies. Schaeffer \emph{et al.} \cite{schaefer2005modeling} discussed the benefits and associated challenges of MDP modeling in the context of medical treatment, and reviewed several instances of medical applications of MDPs, such as spherocytosis treatment and breast cancer screening and treatment.

\begin{figure}[!tb]
\centering
\includegraphics[width=0.46\textwidth]{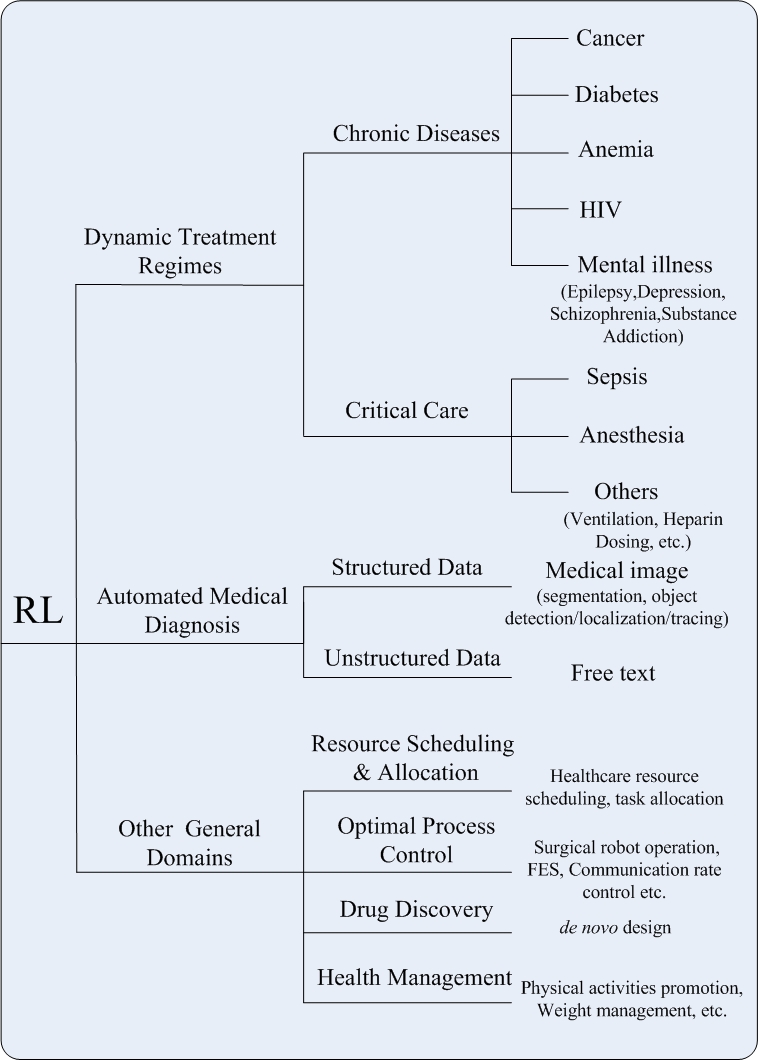}
%\vspace{-0.3cm}
\caption{The outline of application domains of RL in healthcare.}
\label{fig:application_overall_framework}
\end{figure}

With the tremendous theoretical and technical achievements in generalization, representation and efficiency in recent years, RL approaches have been successfully applied in a number of healthcare domains to date. Broadly, these application domains can be categorized into three main types: dynamic treatment regimes in chronic disease or critical care, automated medical diagnosis, and other general domains such as health resources allocation and scheduling, optimal process control, drug discovery and development, as well as health management. Figure \ref{fig:application_overall_framework} provides a diagram outlining the application domains, illustrating how this survey is organized along the lines of the three broad domains in the field.

\section{Dynamic Treatment Regimes}\label{sec:RL_DTR}

One goal of healthcare decision-making is to develop effective treatment regimes that can dynamically adapt to the varying clinical states and improve the long-term benefits of patients. \emph{Dynamic treatment regimes} (DTRs) \cite{chakraborty2014dynamic,laber2014dynamic}, alternatively named as \emph{dynamic treatment policies} \cite{lunceford2002estimation}, \emph{adaptive interventions} \cite{almirall2016adaptive}, or \emph{adaptive treatment strategies} \cite{lavori2008adaptive}, provide a new paradigm to automate the process of developing new effective treatment regimes for individual patients with long-term care \cite{chakraborty2013statistical}. A DTR is composed of a sequence of decision rules to determine the course of actions (e.g., treatment type, drug dosage, or reexamination timing) at a time point according to the current health status and prior treatment history of an individual patient. Unlike traditional randomized controlled trials that are mainly used as an evaluative tool for confirming the efficacy of a newly developed treatment, DTRs are tailored for generating new scientific hypotheses and developing optimal treatments across or within groups of patients \cite{chakraborty2013statistical}. Utilizing valid data generated, for instance, from the \emph{Sequential Multiple Assignment Randomized Trial} (SMART) \cite{murphy2005experimental,murphy2007developing}, an optimal DTR that is capable of optimizing the final clinical outcome of particular interest can be derived.

The design of DTRs can be viewed as a sequential decision making problem that fits into the RL framework well. The series of decision rules in DTRs are equivalent to the policies in RL, while the treatment outcomes are expressed by the reward functions. The inputs in DTRs are a set of clinical observations and assessments of patients, and the outputs are the treatments options at each stage, equivalent to the states and actions in RL, respectively. Apparently, applying RL methods to solve DTR problems demonstrates several benefits. RL is capable of achieving time-dependent decisions on the best treatment for each patient at each decision time, thus accounting for heterogeneity across patients. This precise treatment can be achieved  even without relying on the identification of any accurate mathematical models or explicit relationship between treatments and outcomes. Furthermore, RL driven solutions enable to improve long-term outcomes by considering delayed effect of treatments, which is the major characteristic of medical treatment. Finally, by careful engineering the reward function using expert or domain knowledge, RL provides an elegant way to multi-objective optimization of treatment between efficacy and the raised side effect.

Due to these benefits, RL naturally becomes an appealing tool for constructing optimal DTRs in healthcare. In fact, solving DTR problems accounts for a large proportion of RL studies in healthcare applications, which can be supported by the dominantly large volume of references in this area. The domains of applying RL in DTRs can be classified into two main categories: chronic diseases and critical care.

\subsection{Chronic Diseases}

Chronic diseases are now becoming the most pressing public health issue worldwide, constituting a considerable portion of death every year \cite{world2005preventing}. Chronic diseases normally feature a long period lasting three months or more, expected to require continuous clinical observation and medical care. The widely prevailing chronic diseases include endocrine diseases (e.g.,~diabetes and hyperthyroidism), cardiovascular diseases (e.g.,~heart attacks and hypertension), various mental illnesses (e.g.,~depression and schizophrenia), cancer, HIV infection, obesity, and other oral health problems \cite{chakraborty2013statisticalmethods}. Long-term treatment of these illnesses is often made up of a sequence of medical intervention that must take into account the changing health status of a patient and adverse effects occurring from previous treatment. In general, the relationship of treatment duration, dosage and type against the patient's response is too complex to be be explicitly specified. As such, practitioners usually resort to some protocols following the \emph{Chronic Care Model} (CCM) \cite{wagner2001improving} to facilitate decision making in chronic disease conditions. Since such protocols are derived from average responses to treatment in populations of patients, selecting the best sequence of treatments for an individual patient poses significant challenges due to the diversity across or whithin the population. RL has been utilized to automate the discovery and generation of optimal DTRs in a variety of chronic diseases including caner, diabetes, anemia, HIV and several common mental illnesses.

\subsubsection{Cancer}
\begin{table*}[!tb]
\centering
\caption{Summary of RL Application Examples in the development of DTRs in Cancer}
  \label{tab:cancer}
  \scalebox{0.85}{
    \begin{tabular}{p{2cm}|p{2.6cm}|p{1.6cm}|p{1.3cm}|p{1.7cm}|p{2cm}|p{6.6cm}}
        \toprule
        \hline
        Applications  & References &  Base Methods  &  Efficient Techniques   & Representational Techniques & Data Acquisition & Highlights or Limits \\\hline
          \multirow{8}{2cm}{Optimal chemotherapy drug dosage for cancer treatment}
 &   Zhao \emph{et al.} \cite{zhao2009reinforcement}  & Q-learning &BRL  &N/A& ODE model & Using SVR or ERT to fit Q values; simplistic reward function structure with integer values to assess the tradeoff between efficacy and toxicity.\\ \cline{2-7}
&      Hassani \emph{et al.} \cite{hassani2010reinforcement} & Q-learning & N/A & N/A & ODE model& Naive discrete formulation of states and actions. \\  \cline{2-7}
&    Ahn $\&$ Park \cite{ahn2011drug} &  NAC  & N/A & N/A &ODE model & Discovering the strategy of performing continuous treatment from the beginning. \\ \cline{2-7}
&  Humphrey \cite{humphrey2017using} & Q-learning & BRL  & N/A & ODE model proposed in \cite{zhao2009reinforcement}& Using three machine learning methods to fit Q values, in high dimensional and subgroup scenarios. \\  \cline{2-7}
& Padmanabhan  \cite{padmanabhan2017reinforcement} & Q-learning  & N/A & N/A  &  ODE model&Using different reward functions to model different constraints in cancer treatment.  \\  \cline{2-7}
&    Zhao \emph{et al.} \cite{zhao2011reinforcement} &  Q-learning & BRL  (FQI-SVR)  & N/A & ODE model driven by real NSCLC data& Considering censoring problem in multiple lines of treatment in advanced NSCLC; using overall survival time as the net reward. \\ \cline{2-7}
&     F{\"u}rnkranz \emph{et al.}  \cite{furnkranz2012preference}, Cheng et al. \cite{cheng2011preference} & PI & N/A & PRL &ODE model proposed in \cite{zhao2009reinforcement}& Combining preference learning and RL for optimal therapy design in cancer treatment, but only in model-based DP settings. \\ \cline{2-7}
&    Akrour \emph{et al.} \cite{akrour2012april},  Busa-Fekete et al.  \cite{busa2014preference}  & PS & N/A & PRL &ODE model proposed in \cite{zhao2009reinforcement}& Using active ranking mechanism  to reduce the number of needed ranking queries to the expert to yield a satisfactory policy without a generated model. \\  \hline

 \multirow{4}{2cm}{Optimal fractionation scheduling of radiation therapy for cancer treatment}
&  Vincent \cite{vincent2014reinforcement} & Q-learning, SARSA($\lambda$), TD($\lambda$), PS & BRL (FQI-ERT) & N/A & Linear model, ODE model & Extended ODE model for radiation therapy; using hard constraints in the reward function and simple exploration strategy.\\ \cline{2-7}
& Tseng \emph{et al.} \cite{tseng2017deep} &Q-learning  &  N/A & DRL (DQN) & Data from  114 NSCLC patients &Addressing limited sample size problem using GAN and approximating the transition probability using DNN.  \\ \cline{2-7}
& Jalalimanesh \emph{et al.}\cite{jalalimanesh2017simulation} &  Q-learning &N/A & N/A & Agent-based model & Using agent-based simulation to model the dynamics of tumor growth. \\ \cline{2-7}
 & Jalalimanesh  \emph{et al.}\cite{jalalimanesh2017multi} & Q-learning& N/A & MORL &Agent-based model & Formulated as a multi-objective problem by considering conflicting objective of minimising tumour therapy period and unavoidable side effects.  \\  \hline

\multirow{2}{2cm}{Hypothetical or generic cancer clinical trial}
& Goldberg $\&$ Kosorok  \cite{goldberg2012q},  Soliman \cite{soliman2014personalized} & Q-learning  & N/A& N/A &Linear model& Addressing problems with censored data and a flexible number of stages. \\ \cline{2-7}
& Yauney $\&$ Shah \cite{yauney2018reinforcement} &Q-learning &  N/A & DRL (DDQN) &ODE model& Addressing the problem of unstructured outcome rewards using action-driven rewards. \\ \hline
%&Chu et al. \cite{chu2016adaptive} &29.18 & 7.16 & 118.95 &real datasets from BCSC and WBC& adaptive online learning framework for supporting clinical breast cancer (BC) diagnosis \\ \hline
         \bottomrule
     \end{tabular}
     }
\end{table*}

Cancer is one of the main chronic diseases that causes death. About 90.5 million people had cancer in 2015 and approximately 14 million new cases are occurring each year, causing about 8.8 million annual deaths that account for 15.7\% of total deaths worldwide \cite{stewart2017world}. The primary treatment options for cancer include surgery, chemotherapy, and radiation therapy. To analyze the dynamics between tumor and immune systems, numerous computational models for spatio-temporal or non-spatial tumor-immune dynamics have been proposed and analyzed by researchers over the past decades \cite{eftimie2011interactions}. Building on these models, control policies have been put forward to obtain efficient drug administration (see \cite{ahn2011drug,shi2014survey} and references therein).

Being a sequential evolutionary process by nature, cancer treatment is a major objective of RL in DTR applications \cite{beerenwinkel2014cancer,tenenbaum2010personalizing}. Table \ref{tab:cancer} summaries the major studies of applying RL in various aspects of cancer treatment, from the perspectives of application scenarios (chemotherapy, radiotherapy or generic cancer treatment simulation), basic RL methods, the efficient and representational techniques applied (if applicable), the learning data (retrospective clinical data, or generated from simulation models or computational models), and the main highlights and limits of the study.

RL methods have been extensively studied in deriving efficient treatment strategies for cancer chemotherapy. Zhao \emph{et al.} \cite{zhao2009reinforcement} first applied model-free TD method, Q-learning, for decision making of agent dosage in chemotherapy. Drawing on the chemotherapy mathematical model expressed by several \emph{Ordinary Difference Equations} (ODE),  virtual clinical trial data from \emph{in vivo} tumor growth patterns was quantitatively generated. Two explicit machine learning approaches, \emph{support vector regression} (SVG) \cite{vapnik1997support} and \emph{extremely randomized trees} (ERT) \cite{ernst2005tree}, were applied to fit the approximated Q-functions to the generated trial data. Using this kind of batch learning methods, it was demonstrated that optimal strategies could be extracted directly from clinical trial data in simulation. Ahn and Park \cite{ahn2011drug} studied the applicability of the \emph{Natural AC} (NAC) approach \cite{peters2008natural} to the drug scheduling of cancer chemotherapy based on an ODE-based tumor growth model proposed by de Pillis and Radunskaya \cite{de2003dynamics}. Targeting at minimizing the tumor cell population and the drug amount while maximizing the populations of normal and immune cells, the NAC approach could discover an effective drug scheduling policy by injecting drug continuously from the beginning until an appropriate time. This policy showed better performance than traditional pulsed chemotherapy protocol that administers the drug in a periodical manner, typically on an order of several hours. The superiority of using continuous dosing treatment over a burst of dosing treatment was also supported by the work \cite{hassani2010reinforcement}, where naive discrete Q-learning was applied. More recently, Padmanabhan \emph{et al.} \cite{padmanabhan2017reinforcement} proposed different formulations of reward function in Q-learning to generate effective drug dosing policies for patient groups with different characteristics. Humphrey \cite{humphrey2017using} investigated several supervised learning approaches (\emph{Classification And Regression Trees} (CART), random forests, and modified version of \emph{Multivariate Adaptive Regression Splines} (MARS)) to estimate Q values in a simulation of an advanced generic cancer trial.

Radiotherapy is another major option of treating cancer, and a number of studies have applied RL approaches for developing automated radiation adaptation protocols \cite{feng2018machine}. Jalalimanesh \emph{et al.} \cite{jalalimanesh2017simulation} proposed an agent-based simulation model and Q-learning algorithm to optimize dose calculation in radiotherapy by varying the fraction size during the treatment. Vincent \cite{vincent2014reinforcement} described preliminary efforts in investigating a variety of RL methods to find optimal scheduling algorithms for radiation therapy, including the exhaustive PS \cite{kober2009policy}, FQI \cite{riedmiller2005neural}, SARSA($\lambda$) \cite{rummery1994line} and K-Nearest Neighbors-TD($\lambda$) \cite{de2011robust}. The preliminary findings suggest that there may be an advantage in using non-uniform fractionation schedules for some tissue types.

As the goal of radiotherapy is in essence a multi-objective problem to erase the tumour with radiation while not impacting normal cells as much as possible, Jalalimanesh \emph{et al.} \cite{jalalimanesh2017multi}  proposed a multi-objective distributed Q-learning algorithm to find the Pareto-optimal solutions for calculating radiotherapy dose. Each objective was optimized by an individual learning agent and all the agents compromised their individual solutions in order to derive a Pareto-optimal solution. Under the multi-objective formulation, three different clinical behaviors could be properly modeled (i.e., aggressive, conservative or moderate), by paying different degree of attention to eliminating cancer cells or taking care of normal cells.

A recent study \cite{tseng2017deep} proposed a multi-component DRL framework to automate adaptive radiotherapy decision making for \emph{non-small cell lung cancer} (NSCLC) patients. Aiming at reproducing or mimicking the decisions that have been previously made by clinicians, three neural network components, namely \emph{Generative Adversarial Net} (GAN), transition \emph{Deep Neural Networks} (DNN) and \emph{Deep Q Network} (DQN), were applied: the GAN component was used to generate sufficiently large synthetic patient data from historical small-sized real clinical data; the transition DNN component was employed to learn how states would transit under different actions of dose fractions, based on the data synthesized from the GAN and available real clinical data; once the whole MDP model has been provided, the DQN component was then responsible for mapping the state into possible dose strategies, in order to optimize future radiotherapy outcomes. The whole framework was evaluated in a retrospective dataset of 114 NSCLC patients who received radiotherapy under a successful dose escalation protocol. It was demonstrated that the DRL framework was able to learn effective dose adaptation policies between 1.5 and 3.8 Gy, which complied with the original dose range used by the clinicians.

% censoring
The treatment of cancer poses several significant theoretical problems for applying existing RL approaches. Patients may drop out the treatment anytime due to various uncontrolled reasons, causing the final treatment outcome (e.g., survival time in cancer treatment) unobserved. This \emph{data censoring problem} \cite{goldberg2012q} complicates the practical use of RL in discovering individualized optimal regimens. Moreover, in general cancer treatment, the initiation and timing of the next line of therapy depend on the disease progression, and thus the number of treatment stage can be flexible. For instance, NSCLC patients usually receive one to three treatment lines, and the necessity and timing of the second and third lines of treatment vary from person to person. Developing valid methodology for computing optimal DTRs in such a flexible setting is currently a premier challenge. Zhao \emph{et al.} \cite{zhao2011reinforcement} presented an adaptive Q-learning approach to discover optimal DTRs for the first and second lines of treatment in Stage IIIB/IV NSCLC. The trial was conducted by randomizing the different compounds for first and second-line treatments, as well as the timing of initiating the second-line therapy. In order to successfully handle the complex censored survival data, a modification of SVG approach, $\epsilon-SVR-C$, was proposed to estimate the optimal Q values. A simulation study showed that the approach could select optimal compounds for two lines of treatment directly from clinical data, and the best initial time for second-line therapy could be derived while taking into account the heterogeneity across patients. Other studies \cite{goldberg2012q,soliman2014personalized}  presented the novel \emph{censored-Q-learning} algorithm that is adjusted for a multi-stage decision problem with a flexible number of stages in which the rewards are survival times that are subject to censoring.

%preference
To tackle the problem that a numerical reward function should be specified beforehand in standard RL techniques, several studies investigated the possibility of formulating rewards using qualitative preference or simply based on past actions in the treatment of cancer \cite{cheng2011preference,furnkranz2012preference,yauney2018reinforcement}. Akrour \emph{et al.} \cite{akrour2012april} proposed a PRL method combined with active ranking in order to decrease the number of ranking queries to the expert needed to yield a satisfactory policy. Experiments on the cancer treatment testbeds showed that a very limited external information in terms of expert's ranking feedbacks might be sufficient to reach state-of-the-art results. Busa-Fekete \emph{et al.} \cite{busa2014preference} introduced a preference-based variant of a direct PS method in the medical treatment design for cancer clinical trials. A novel approach based on action-driven rewards was first proposed in \cite{yauney2018reinforcement}. It was showed that new dosing regimes in cancer chemotherapy could be learned using action-derived penalties, suggesting the possibility of using RL methods in situations when final outcomes are not available, but priors on beneficial actions can be more easily specified.

\subsubsection{Diabetes}
\emph{Diabetes mellitus}, or simply called \emph{diabetes}, is one of the most serious chronic diseases in the world. According to a recent report released by \emph{International Diabetes Federation} (IDF), there are 451 million people living with diabetes in 2017, causing approximately 5 million deaths worldwide and USD 850 billion global healthcare expenditure \cite{cho2018idf}. It is expected that by 2045, the total number of adults with diabetes would increase to near 700 million, accounting for 9.9\% of the adult population. Since the high prevalence of diabetes presents significant social influence and financial burdens, there has been an increasing urgency to ensure effective treatment to diabetes across the world.

Intensive research concern has been devoted to the development of effective blood glucose control strategies in treatment of insulin-dependent diabetes (i.e., type 1 diabetes). Since its first proposal in the 1970s \cite{albisser1974clinical},  \emph{artificial pancreas} (AP) have been widely used in the blood glucose control process to compute and administrate a precise insulin dose, by using a \emph{continuous glucose monitoring system} (CGMS) and a closed-loop controller \cite{cobelli2011artificial}. Tremendous progress has been made towards insulin infusion rate automation in AP using traditional control strategies such as \emph{Proportional-Integral-Derivative} (PID), \emph{Model Predictive Control} (MPC), and \emph{Fuzzy Logic} (FL) \cite{bequette2005critical,peyser2014artificial}. A major concern is the inter- and intra- variability of the diabetic population which raises the demand for a personalized, patient specific approach of the glucose regulation. Moreover, the complexity of the physiological system, the variety of disturbances such as meal, exercise, stress and sickness, along with the difficulty in modelling accurately the glucose-insulin regulation system all raise the need in the development of more advanced adaptive algorithms for the glucose regulation.

RL approaches have attracted increasingly high attention in personalized, patient specific glucose regulation in AP systems \cite{bothe2013use}. Yasini \emph{et al.} \cite{yasini2009agent} made an initial study on using RL to control an AP to maintain normoglycemic around 80 mg/dl. Specifically, model-free TD Q-learning algorithm was applied to compute the insulin delivery rate, without relying on an explicit model of the glucose-insulin dynamics. Daskalaki \emph{et al.} \cite{daskalaki2010preliminary} presented an AC controller for the estimation of insulin infusion rate \emph{in silico} trial based on the University of Virginia/Padova type 1 diabetes simulator \cite{kovatchev2009silico}. In an evaluation of 12 day meal scenario for 10 adults, results showed that the approach could prevent hypoglycaemia well, but hyperglycaemia could not be properly solved due to the static behaviors of the \emph{Actor} component. The authors then proposed using daily updates of the average \emph{basal rate} (BR) and the\emph{ insulin-to-carbohydrate} (IC) ratio in order to optimize glucose regulation  \cite{daskalaki2013actor},  and using estimation of \emph{information transfer} (IT) from insulin to glucose for automatic and personalized tuning of the AC  approach \cite{daskalaki2013personalized}. This idea was motivated by the fact that small adaptation of insulin in the \emph{Actor} component may be sufficient in case of large amount of IT from insulin to glucose, whereas more dramatic updates may be required for low IT. The results from the \emph{Control Variability Grid Analysis} (CVGA) showed that the approach could achieve higher performance in all three groups of patients, with 100\% percentages in the A+B zones for adults, and 93\% for both adolescents and children, compared to approaches with random initialization and zero initial values. The AC approach was significantly extended to directly link to patient-specific characteristics, and evaluated more extensively under a complex meal protocol, meal uncertainty and insulin sensitivity variation \cite{daskalaki2016model,sun2018dual}.

A number of studies used certain mathematical models to simulate the glucose-insulin dynamic system in patients. Based on the Palumbo mathematical model \cite{palumbo2007qualitative}, the on-policy SARSA was used for insulin delivery rate \cite{noori2017glucose}.  Ngo \emph{et al.}  applied model-based VI method \cite{ngo2018reinforcement} and AC method \cite{ngo2018control} to reduce the fluctuation of the blood glucose in both fasting and post-meal scenarios, drawing on the Bergman's minimal insulin-glucose kinetics model \cite{bergman1979quantitative} and the Hovorka model \cite{hovorka2004nonlinear} to simulate a patient. De Paula \emph{et al.} \cite{de2015controlling,de2015line} proposed policy learning algorithms that integrates RL with Gaussian processes to take into account glycemic variability under uncertainty, using the Ito's stochastic model of the glucose-insulin dynamics \cite{acikgoz2010blood}.

There are also several data-driven studies carried out to analyze RL in diabetes treatment based on real data from diabetes patients. Utilizing the data extracted from the medical records of over 10,000 patients in the University of Tokyo Hospital, Asoh \emph{et al.} \cite{asoh2013modeling} estimated the MDP model underlying the progression of patient state and evaluated the value of treatment using the VI method. The opinions of a doctor were used to define the reward for each treatment. The preassumption of this predefined reward function then motivated the application of IRL approach to reveal the reward function that doctors were using during their treatments \cite{asoh2013application}. Using observational data on the effect of food intake and physical activity in an outpatient setting using mobile technology, Luckett \emph{et al.} \cite{luckett2018estimating} proposed the \emph{V-learning} method that directly estimates a policy which maximizes the value over a class of policies and requires minimal assumptions on the data-generating process. The method has been applied to estimate treatment regimes to reduce the number of hypo and hyperglycemic episodes in patients with type 1 diabetes.

\subsubsection{Anemia}
Anemia is a common comorbidity in chronic renal failure that occurs in more than 90\% of patients with \emph{end-stage renal disease} (ESRD) who are undertaking hemodialysis. Caused by a failure of adequately producing \emph{endogenous erythropoietin} (EPO) and thus red blood cells, anemia can have significant impact on organ functions, giving rise to a number of severe consequences such as heart disease or even increased mortality. Currently, anemia can be successfully treated by administering \emph{erythropoiesis-stimulating agents} (ESAs), in order to maintain the \emph{hemoglobin} (HGB) level within a narrow range of 11-12 g/dL. To achieve this, professional clinicians must carry out a labor intensive process of dosing ESAs to assess monthly HGB and iron levels before making adjustments accordingly. However, since the existing \emph{Anemia Management Protocol} (AMP) does not account for the high inter- and intra-individual variability in the patient's response, the HGB level of some patients usually oscillates around the target range, causing several risks and side-effects.

As early as in 2005, Gaweda \emph{et al.} \cite{gaweda2005reinforcement} first proposed using RL to perform individualized treatment in the management of renal anemia. The target under control is the HGB, whereas the control input is the amount of EPO administered by the physician. As the iron storage in the patient, determined by \emph{Transferrin Saturation} (TSAT), also has an impact on the process of red blood cell creation, it is considered as a state component together with HGB. To model distinct dose-response relationship within a patient population, a fuzzy model was estimated first by using real records of 186 hemodialysis patients from the Division of Nephrology, University of Louisville. On-policy TD method, SARSA, was then performed on the sample trajectories generated by the model. Results show that the proposed approach generates adequate dosing strategies for representative individuals from different response groups. The authors then proposed a combination of MPC approach with SARSA for decision support in anemia management \cite{gaweda2006model}, with the MPC component used for simulation of patient response and SARSA for optimization of the dosing strategy. However, the automated RL approaches in these studies could only achieve a policy with a comparable outcome against the existing AMP. Other studies applied various kinds of Q-learning, such as Q-learning with function approximation, or directly based on state-aggregation \cite{gaweda2005individualization,martin2009reinforcement,martin2007validation}, in providing effective treatment regimes in anemia.

Several studies resorted to BRL methods to derive optimal ESA dosing strategies for anemia treatment. By performing a retrospective study of a cohort of 209 hemodialysis patients, Malof and Gaweda \cite{malof2011optimizing} adopted the batch FQI method to achieve dosing strategies that were superior to a standard AMP. The FQI method was also applied by Escandell \emph{et al.} \cite{escandell2011adaptive} for discovering efficient dosing strategies based on the historical treatment data of 195 patients in nephrology centers allocated around Italy and Portugal. An evaluation of the FQI method on a computational model that describes the effect of ESAs on the hemoglobin level showed that FQI could achieve an increment of 27.6\% in the proportion of patients that are within the targeted range of hemoglobin during the period of treatment. In addition, the quantity of drug needed is reduced by 5.13\%, which indicates a more efficient use of ESAs  \cite{escandell2014optimization}.

\subsubsection{HIV}
Discovering effective treatment strategies for HIV-infected individuals remains one of the most significant challenges in medical research. To date, the effective way to treat HIV makes use of a combination of anti-HIV drugs (i.e., antiretrovirals) in the form of \emph{Highly Active Antiretroviral Therapy} (HAART) to inhibit the development of drug-resistant HIV strains \cite{adams2004dynamic}. Patients suffering from HIV are typically prescribed a series of treatments over time in order to maximize the long-term positive outcomes of reducing patients' treatment burden and improving adherence to medication. However, due to the differences between individuals in their immune responses to treatment, discovering the optimal drug combinations and scheduling strategy is still a difficult task in both medical research and clinical trials.

Ernst \emph{et al.} \cite{ernst2006clinical} first introduced RL techniques in computing \emph{Structured Treatment Interruption} (STI) strategies for HIV infected patients. Using a mathematical model \cite{adams2004dynamic} to artificially generate the clinical data, the BRL method FIQ-ERT was applied to learn an optimal drug prescription strategy in an off-line manner. The derived STI strategy is featured with a cycling between the two main anti-HIV drugs: \emph{Reverse Transcriptase Inhibitors} (RTI) and \emph{Protease Inhibitors} (PI), before bringing the patient to the healthy drug-free steady-state. Using the same mathematical model, Parbhoo \cite{parbhoo2014reinforcement} further implemented three kinds of BRL methods, FQI-ERT, neural FQI and LSPI, to the problem of HIV treatment, indicating that each learning technique had its own advantages and disadvantages. Moreover, a testing based on a ten-year period of real clinical data from 250 HIV-infected patients in Charlotte Maxeke Johannesburg Academic Hospital, South Africa verified that the RL methods were capable of suggesting treatments that were reasonably compliant with those suggested by clinicians.

A mixture-of-experts approach was proposed in \cite{parbhoo2017combining} to combine the strengths of both kernel-based regression methods (i.e., history-alignment model) and RL (i.e., model-based Bayesian PORL) for HIV therapy selection. Since kernel-based regression methods are more suitable for modeling more related patients in history, while model-based RL methods are more suitable for reasoning about the future outcomes, automatically selecting an appropriate model for a particular patient between these two methods thus tends to provide simpler yet more robust patterns of response to the treatment. Making use of a subset of the EuResist database consisting of HIV genotype and treatment response data for 32,960 patients, together with the 312 most common drug combinations in the cohort, the treatment therapy derived by the mixture-of-experts approach outperformed those derived by each method alone.

Since the treatment of HIV highly depends the patient's immune system that varies from person to person, it is thus necessary to derive efficient learning strategies that can address and identify the variations across subpopulations. Marivate \emph{et al.} \cite{marivate2014quantifying} formalized a routine to accommodate multiple sources of uncertainty in BRL methods to better evaluate the effectiveness of treatments across a subpopulations of patients. Other approaches applied various kinds of TRL techniques so as to take advantage of the prior information
from previously learned transition models \cite{killian2016transfer,killian2017robust} or learned policy \cite{yao2018direct}. More recently, Yu \emph{et al.} \cite{yu2019incorporating} proposed a causal policy gradient algorithm and evaluated it in the treatment of HIV in order to facilitate the final learning performance and increase explanations of learned strategies.

The treatment of HIV provides a well-known testbed for evaluation of exploration mechanisms in RL research. Simulations show that the basin of attraction of the healthy steady-state is rather small compared to that of the non-healthy steady state \cite{adams2004dynamic}. Thus, general exploration methods are unable to yield meaningful performance improvement as they can only obtain samples in the vicinity of the ``non-healthy'' steady state. To solve this issue, several studies have proposed more advanced exploration strategies in order to increase the learning performance in HIV treatment. Pazis \emph{et al.} \cite{pazis2013pac} introduced an algorithm for PAC optimal exploration in continuous state spaces. Kawaguchi considered the time bound in a PAC exploration process \cite{kawaguchi2016bounded}. Results in both studies showed that the exploration algorithm could achieve far better strategies than other existing exploration strategies in HIV treatment.

\subsubsection{Mental Disease}
Mental diseases are characterized by a long-term period of clinical treatments that usually require adaptation in the duration, dose, or type of treatment over time \cite{murphy2007methodological}. Given that the brain is a complex system and thus extremely challenging to model, applying traditional control-based methods that rely on accurate brain models in mental disease treatment is proved infeasible. Well suited to the problem at hand, RL has been widely applied to DTRs in a wide range of mental illness including epilepsy, depression, schizophrenia and various kinds of substance addiction.

\paragraph{Epilepsy}
Epilepsy is one of the most common severe neurological disorders, affecting around 1\% of the world population. When happening, epilepsy is manifested in the form of intermittent and intense seizures that are recognized as abnormal synchronized firing of neural populations. Implantable electrical deep-brain stimulation devices are now an important treatment option for drug-resistant epileptic patients. Researchers from nonlinear dynamic systems analysis and control have proposed promising prediction and detection algorithms to suppress the frequency, duration and amplitude of seizures \cite{alotaiby2014eeg}. However, due to lack of full understanding of seizure and its associated neural dynamics, designing optimal seizure suppression algorithms via minimal electrical stimulation has been for a long time a challenging task in treatment of epilepsy.

RL enables direct closed-loop optimizations of deep-brain stimulation strategies by adapting control policies to patients' unique neural dynamics, without necessarily relying on having accurate prediction or detection of seizures. The goal is to explicitly maximize the effectiveness of stimulation, while simultaneously minimizing the overall amount of stimulation applied thus reducing cell damage and preserving cognitive and neurological functions \cite{panuccio2018progress}. Guez \emph{et al.} \cite{guez2008adaptive,pineau2009treating,guez2010adaptive} applied the BRL method, FQI-ERT, to optimize a deep-brain stimulation strategy for the treatment of epilepsy. Encoding the observed \emph{Electroencephalograph} (EEG) signal as a 114-dimensional continuous feature vector, and four different simulation frequencies as the actions, the RL approach was applied to learn an optimal stimulation policy using data from an \emph{in vitro} animal model of epilepsy (i.e., field potential recordings of seizure-like activity in slices of rat brains). Results showed that RL strategies substantially outperformed the current best stimulation strategies in the literature, reducing the incidence of seizures by 25\% and total amount of electrical stimulation to the brain by a factor of about 10. Subsequent validation work \cite{panuccio2013adaptive} showed generally similar results that RL-based policy could prevent epilepsy with a significant reduced amount of stimulation, compared to fixed-frequency stimulation strategies. Bush and Pineau \cite{bush2009manifold} applied manifold embeddings to reconstruct the observable state space in MRL, and applied the proposed approach to tackle the high complexity of nonlinearity and partially observability in real-life systems. The learned neurostimulation policy was evaluated to suppress epileptic seizures on animal brain slices and results showed that seizures could be effectively suppressed after a short transient period.

While the above \emph{in vitro} biological models of epilepsy are useful for research, they are nonetheless time-consuming and associated with high cost. In contrast, computational models can provide large amounts of reproducible and cheap data that may permit precise manipulations and deeper investigations. Vincent \cite{vincent2014reinforcement} proposed an \emph{in silico} computational model of epileptiform behavior in brain slices, which was verified by using biological data from rat brain slices \emph{in vitro}. Nagaraj \emph{et al.} \cite{nagaraj2017seizure} proposed the first computational model that captures the transition from inter-ictal to ictal activity, and applied naive Q-learning method to optimize stimulation frequency for controlling seizures with minimum stimulations. It was shown that even such simple RL methods could converge on the optimal solution in simulation with slow and fast inter-seizure intervals.

\paragraph{Depression}
\emph{Major depressive disorder} (MDD), also known simply as depression, is a mental disorder characterized by at least two weeks of low mood that is present across most situations.  Using data from the \emph{Sequenced Treatment Alternatives to Relieve Depression} (STAR*D) trial \cite{rush2004sequenced}, which is a sequenced four-stage randomized clinical trial of patients with MDD, Pineau \emph{et al.} \cite{pineau2007constructing} first applied Kernel-based BRL \cite{ormoneit2002kernel} for constructing useful DTRs for patients with MDD. Other work tries to address the problem of nonsmooth of decision rules as well as nonregularity of the parameter estimations in traditional RL methods by proposing various extensions over default Q-learning procedure in order to increase the robustness of learning \cite{chakraborty2013inference}. Laber \emph{et al.} \cite{laber2014interactive} proposed a new version of Q-learning, \emph{interactive Q-learning} (IQ-learning), by interchanging the order of certain steps in traditional Q-learning, and showed that IQ-learning improved on Q-learning in terms of integrated mean squared error in a study of MDD. The IQ-learning framework was then extended to optimize functionals of the outcome distribution other than the expected value \cite{linn2014interactive,linn2017interactive}. Schulte \emph{et al.} \cite{schulte2014q} provided systematic empirical studies of Q-learning and \emph{Advantage-learning} (A-learning) \cite{murphy2003optimal} methods and illustrated their performance using data from an MDD study. Other approaches include the \emph{penalized Q-learning} \cite{song2015penalized}, the \emph{Augmented Multistage Outcome-Weighted Learning} (AMOL) \cite{liu2016robust}, the budgeted learning algorithm \cite{deng2014budgeted}, and the \emph{Censored Q-learning} algorithm \cite{soliman2014personalized}.

\paragraph{Schizophrenia}
RL methods have been also used to derive optimal DTRs in treatment of schizophrenia, using data from the \emph{Clinical Antipsychotic Trials of Intervention Effectiveness} (CATIE) study \cite{keefe2007neurocognitive}, which was an 18-month study divided into two main phases of treatment. An in-depth case study of using BRL, FQI, to optimize treatment choices for patients with schizophrenia using data from CATIE was given by \cite{shortreed2011informing}. Key technical challenges of applying RL in typically continuous, highly variable, and high-dimensional clinical trials with missing data were outlined. To address these issues, the authors proposed the use of multiple imputation to overcome the missing data problem, and then presented two methods, \emph{bootstrap voting} and \emph{adaptive confidence intervals}, for quantifying the evidence in the data for the choices made by the learned optimal policy. Ertefaie \emph{et al.} \cite{ertefaie2016q} accommodated residual analyses into Q-learning in order to increase the accuracy of model fit and demonstrated its superiority over standard Q-learning using data from CATIE.

Some studies have focused on optimizing multiple treatment objectives in dealing with schizophrenia. Lizotte \emph{et al.} \cite{lizotte2012linear} extended the FQI algorithm by considering multiple rewards of symptom reduction, side-effects and quality of life simultaneously in sequential treatments for schizophrenia. However, it was assumed that end-users had a true reward function that was linear in the objectives and all future actions could be chosen optimally with respect to the same true reward function over time. To solve these issues, the authors then proposed the non-deterministic multi-objective FIQ algorithm, which computed policies for all preference functions simultaneously from continuous-state, finite-horizon data \cite{lizotte2016multi}.
When patients do not know or cannot communicate their preferences, and there is heterogeneity across patient preferences for these outcomes, formation of a single composite outcome that correctly balances the competing outcomes for all patients is not possible. Laber \emph{et al.} \cite{laber2014set} then proposed a method for constructing DTRs for schizophrenia that accommodates competing outcomes and preference heterogeneity across both patients and time by recommending sets of treatments at each decision point. Butler \emph{et al.} \cite{butler2017incorporating} derived a preference sensitive optimal DTR for schizophrenia patient by directly eliciting patients' preferences overtime.

\paragraph{Substance Addiction}
\emph{Substance addiction}, or \emph{substance use disorder} (SUD), often involves a chronic course of repeated cycles of cessation followed by relapse \cite{dennis2007managing,almirall2016adaptive}. There has been great interest in the development of DTRs by investigators to deliver in-time interventions or preventions to end-users using RL methods, guiding them to lead healthier lives. For example, Murphy \emph{et al.} \cite{murphy2016batch} applied AC algorithm to reduce heavy drinking and smoking for university students. Chakraborty \emph{et al.} \cite{chakraborty2013statistical,chakraborty2010inference,chakraborty2008bias} used Q-learning with linear models to identify DTRs for smoking cessation treatment regimes. Tao \emph{et al.} \cite{tao2018tree} proposed a tree-based RL method to directly estimate optimal DTRs, and identify dynamic SUD treatment regimes for adolescents.

\subsection{Critical Care}
Unlike the treatment of chronic diseases, which usually requires a long period of constant monitoring and medication, critical care is dedicated to more seriously ill or injured patients that are in need of special medical treatments and nursing care. Usually, such patients are provided with separate geographical area, or formally named the \emph{intensive care unit} (ICU), for intensive monitoring and close attention, so as to improve the treatment outcomes \cite{vincent2013critical}. ICUs will play a major role in the new era of healthcare systems. It is estimated that the ratio of ICU beds to hospital beds would increase from 3-5\% in the past to 20-30\% in the future \cite{krell2008critical}.

Significant attempts have been devoted to the development of clearer guidelines and standardizing approaches to various aspects of interventions in ICUs, such as sedation, nutrition, administration of blood products, fluid and vasoactive drug therapy, haemodynamic endpoints, glucose control, and mechanical ventilation \cite{vincent2013critical}. Unfortunately, only a few of these interventions could be supported by high quality evidence from randomised controlled trials or meta-analyses \cite{ghassemi2015state}, especially when it comes to development of potentially new therapies for complex ICU syndromes, such as sepsis \cite{rhodes2017surviving} and acute respiratory distress syndrome \cite{force2012acute}.

Thanks to the development in ubiquitous monitoring and censoring techniques, it is now possible to generate rich ICU data in a variety of formats such as free-text clinical notes, images, physiological waveforms, and vital sign time series, suggesting a great deal of opportunities for the applications of machine learning and particularly RL techniques in critical care \cite{kamio2017use,vellido2018machine}. However, the inherent 3C (\emph{Compartmentalization}, \emph{Corruption}, and \emph{Complexity}) features indicate that critical care data are usually noisy, biased and incomplete \cite{johnson2016machine}. Properly processing and interpreting this data in a way that can be used by existing machine learning methods is the premier challenge of data analysis in critical care. To date, RL has been widely applied in the treatment of sepsis (Section \ref{sec:sepsis}), regulation of sedation (Section \ref{sec:anesthesia}), and some other decision making problems in ICUs  such as mechanical ventilation and heparin dosing (Section \ref{sec:ICU_other}). Table \ref{tab:critical_care} summarizes these applications according to the applied RL techniques and the sources of data acquired during learning.

\begin{table*}[!tb]
\centering
\caption{Summary of RL Application Examples in the Development of DTRs in Critical Care}
  \label{tab:critical_care}
  \scalebox{0.75}{
    \begin{tabular}{p{1.2cm}|p{3cm}|p{3cm}|p{1.7cm}|p{1.3cm}|p{2.0cm}|p{2cm}|p{6.4cm}}
        \toprule
        \hline
         Domain& Application&Reference &  Base method  &  Efficient Techniques   & Representational Techniques & Data Acquisition & Highlights and Limits \\\hline
        \multirow{5}{*}{Sepsis}&
        \multirow{3}{3cm}{Administration of IV fluid and maximum VP}
 & Komorowski \emph{et al.} \cite{komorowski2016markov,komorowski2018artificial}  & SARSA,PI & N/A & N/A & MIMIC-III &Naive application of SARSA and PI in a discrete state and action-space. \\ \cline{3-8}
 && Raghu \emph{et al.} \cite{raghu2017deep,raghu2017continuous} & Q-learning & N/A &DRL (DDDQN)&MIMIC-III&Application of DRL in a fully continuous state but discrete action space.\\ \cline{3-8}
  &&  Raghu \emph{et al.} \cite{raghu2018model}  & PS & MRL &N/A&MIMIC-III& Model-based learning with continuous state-space; integrating clinician's policies into RL policies.\\ \cline{3-8}
    &&    Utomo \emph{et al.}   \cite{utomo2018treatment}  & MC & N/A &N/A&MIMIC-III& Estimating transitions of patient health conditions and treatments to increase its explainability.\\ \cline{3-8}
 &&  Peng \emph{et al.} \cite{peng2019improving} & Q-learning & N/A &DRL (DDDQN) &MIMIC-III&Adaptive switching between kernel learning and DRL. \\ \cline{3-8}
 &&  Futoma \emph{et al.} \cite{futoma2018learning} & Q-learning &N/A &DRL& Clinical data at university hospital& Tackling sparsely sampled and frequently missing multivariate time series data.\\ \cline{3-8}
  &&  Yu \emph{et al.} \cite{yu2019deep} & Q-learning &BRL(FQI) &DRL, IRL& MIMIC-III& Inferring the best reward functions using deep IRL.\\ \cline{3-8}
 & & Li \emph{et al.} \cite{li2018actor} & AC &N/A &PORL&MIMIC-III&Taking into account uncertainty and history information of sepsis patients.\\ \cline{2-8}
     &    \multirow{1}{3cm}{Targeted blood glucose regulation}
   &  Weng \emph{et al.} \cite{weng2017representation}  & PI &N/A &N/A&MIMIC-III&Learning the optimal targeted blood glucose levels for sepsis patients\\ \cline{2-8}
    &      \multirow{1}{3cm}{Cytokine mediation}
 &Petersen \emph{et al.} \cite{petersen2018precision}  & AC &N/A &DRL (DDPG)& Agent-based model &  Using reward shaping to facilitate the learning efficiency; significantly reducing mortality from $49\%$ to $0.8\%$. \\ \hline

         \multirow{8}{*}{Anesthesia}& \multirow{8}{3cm}{Regulation and automation of sedation and analgesia to maintain physiological stability and lowering pains of patients}
 & Moore \emph{et al.} \cite{moore2004intelligent,sinzinger2005sedation} & $Q(\lambda)$ &N/A &N/A& PK/PD model &Achieving superior stability compared to a well-tuned PID controller.\\ \cline{3-8}
& &  Moore \emph{et al.} \cite{moore2011reinforcement1,moore2011reinforcement2} &Q-learning  &N/A &N/A& PK/PD model&Using the change of BIS as the state representation. \\ \cline{3-8}
&  &Moore \emph{et al.} \cite{moore2014reinforcement,moore2010reinforcement} & Q-learning  &N/A  &N/A & \emph{In vivo} study &First clinical trial for anesthesia administration using RL on human volunteers.\\ \cline{3-8}
 &  & Sadati \emph{et al.} \cite{sadati2006multivariable} & Unclear &  N/A &N/A&PK/PD model &Expert knowledge can be used to realize reasonable initial dosage and keep drug inputs in safe values.\\ \cline{3-8}
  &    &Borera \emph{et al.} \cite{borera2011adaptive} & Q-learning & N/A &N/A &PK/PD model& Using an adaptive filter to eliminate the delays when estimating patient state.\\ \cline{3-8}
   &      & Lowery $\&$ Faisal \cite{lowery2013towards}& AC &N/A & N/A& PK/PD model&Considering the continuous state and action spaces. \\ \cline{3-8}
    &        & Padmanabhan \emph{et al.} \cite{padmanabhan2015closed} & Q-learning & N/A & N/A& PK/PD model&Regulating sedation and hemodynamic parameters simultaneously.\\ \cline{3-8}
 & &Humbert \emph{et al.} \cite{humbertlearning}  & N/A & N/A &POMDP, IRL&Clinical data& Training an RL agent to mimic decisions by expert anesthesiologists.\\ \hline

 \multirow{4}{*}{Others}
 &   \multirow{2}{*}{Heparin Dosing}
&  Nemati \emph{et al.} \cite{nemati2016optimal} & Q-learning & BRL &PORL&MIMIC II& End-to-end learning with hidden states of patients.\\ \cline{3-8}
 && Lin \emph{et al.} \cite{lin2018deep} &AC & N/A&DRL(DDPG)&MIMIC, Emory Healthcare data& Addressing dosing problems in continuous state-action spaces.\\ \cline{2-8}
&   \multirow{1}{3cm}{General medication recommendation}
&  Wang \emph{et al.} \cite{wang2018supervised}  & AC  & N/A &DRL (DDPG)&MIMIC-III& Combining supervised and reinforcement learning for medication dosing covering a large number of diseases. \\ \cline{2-8}
&   \multirow{2}{3cm}{Mechanical ventilation and sedative dosing}
& Prasad \emph{et al.} \cite{prasad2017reinforcement} & Q-learning  & BRL(FQI) & N/A &MIMIC-III&Optimal decision making for the weaning time of mechanical ventilation and personalized sedation dosage.\\ \cline{3-8}
& &Yu \emph{et al.} \cite{yu2019inverse} & Q-learning  & BRL(FQI) & IRL &MIMIC-III&Applying IRL in inferring the reward functions.\\ \cline{3-8}
& &Yu \emph{et al.} \cite{yu2020supervised} & AC  & N/A & N/A &MIMIC-III&Combing supervised learning and AC for more efficient decision making.\\ \cline{3-8}
&& Jagannatha \emph{et al.} \cite{jagannathatowards} & Q-learning, PS  & BRL(FQI) & N/A&MIMIC-III&Analyzing limitations of off-policy policy evaluation methods in ICU settings.\\ \cline{2-8}
&  \multirow{2}{3cm}{Ordering of lab tests }
 &Cheng \emph{et al.} \cite{cheng2018optimal}& Q-learning  & BRL(FQI)  & MORL&MIMIC III&Designing a multi-objective reward function that reflects clinical considerations when ordering labs. \\ \cline{3-8}
&  &Chang et al. \cite{chang2018dynamic}& Q-learning & N/A  & DRL (Dueling DQN)&MIMIC III&The first RL application on multi-measurement scheduling problem in the clinical setting.\\\cline{2-8}
 &  \multirow{1}{3cm}{Prevention and treatments for GVHD}
 & Krakow \emph{et al.} \cite{krakow2017tools}& Q-learning  & N/A  & N/A&CIBMTR data&First proposal of DTRs for acute GVHD prophylaxis and treatment.\\ \cline{3-8}
 & &  Liu \emph{et al.} \cite{liu2017deep}&  Q-learning & N/A  & DRL (DQN)&CIBMTR data& Incorporation of a supervised learning step into RL.\\ \hline
         \bottomrule
     \end{tabular}
     }
\end{table*}

\subsubsection{Sepsis} \label{sec:sepsis}
Sepsis, which is defined as severe infection causing life-threatening acute organ failure, is a leading cause of mortality and associated healthcare costs in critical care \cite{gotts2016sepsis}. While numbers of international organizations have devoted significant efforts to provide general guidance for treating sepsis over the past 20 years, physicians at practice still lack universally agreed-upon decision support for sepsis \cite{rhodes2017surviving}. With the available data obtained from freely accessible critical care databases such as the \emph{Multiparameter Intelligent Monitoring in Intensive Care} (MIMIC) \cite{johnson2016mimic}, recent years have seen an increasing number of studies that applied RL techniques to the problem of deducing optimal treatment policies for patients with sepsis \cite{saria2018individualized}.

The administration of \emph{intravenous} (IV)  and maximum \emph{vasopressor} (VP) is a key research and clinical challenge in sepsis. A number of studies have been carried out to tackle this issue  in the past years. Komorowski \emph{et al.} \cite{komorowski2016markov,komorowski2018artificial} directly applied the on-policy SARSA algorithm and model-based PI method in a discretized state and action-space. Raghu \emph{et al.} \cite{raghu2017deep,raghu2017continuous} examined fully continuous state and action space, where policies are learned directly from the physiological state data. To this end, the authors proposed the fully-connected \emph{Dueling Double DQN} to learn an approximation for the optimal action-value function, which combines three state-of-the-art efficiency and stability boosting techniques in DRL, i.e., \emph{Double DQN} \cite{van2016deep}, \emph{Dueling DQN} \cite{wang2016dueling} and \emph{Prioritized Experience Replay} (PER) \cite{schaul2015prioritized}. Experimental results demonstrated that using continuous state-space modeling could identify interpretable policies with improved patient outcomes, potentially reducing patient mortality in the hospital by 1.8 - 3.6\%. The authors also directly estimated the transition model in continuous state-space, and applied two PS methods, the direct policy gradient and \emph{Proximal Policy Optimization} PPO \cite{schulman2017proximal}, to derive a treatment strategy \cite{raghu2018model}.  Utomo \emph{et al.}   \cite{utomo2018treatment} proposed a graphical model that was able to show transitions of patient health conditions and treatments for better explanability, and applied MC to generate a real-time treatment recommendation. Li \emph{et al.} \cite{li2018actor} provided an online POMDP solution to take into account uncertainty and history information in sepsis clinical applications.  Futoma \emph{et al.} \cite{futoma2018learning} used multi-output Gaussian processes and DRL to directly learn from sparsely sampled and frequently missing multivariate time series ICU data.  Peng \emph{et al.} \cite{peng2019improving} applied the mixture-of-experts framework \cite{parbhoo2017combining} in sepsis treatment by automatically switching between kernel learning and DRL depending on patient's current history. Results showed that this kind of mixed learning could achieve better performance than the strategies by physicians, Kernel learning and DQN learning alone. Most recently, Yu \emph{et al.} \cite{yu2019deep} addressed IRL problems in sepsis treatment.

Targeting at glycemic regulation problems for severely ill septic patients, Weng \emph{et al.} \cite{weng2017representation} applied PI to learn the optimal targeted blood glucose levels from real data trajectories. Petersen \emph{et al.} \cite{petersen2018precision} investigated the cytokine mediation problem in sepsis treatment, using the DRL method, \emph{Deep Deterministic Policy Gradient} (DDPG) \cite{lillicrap2015continuous}, to tackle the hi-dimensional continuous states and actions, and potential-based reward shaping \cite{ng1999policy} to facilitate the learning efficiency. The proposed approach was evaluated using an agent-based model, the \emph{Innate Immune Response Agent-Based Model} (IIRABM), that simulates the immune response to infection. The learned treatment strategy was showed to achieve 0.8\% mortality over 500 randomly selected patient parameterizations with mortalities average of 49\%, suggesting that adaptive, personalized multi-cytokine mediation therapy could be promising for treating sepsis.

\subsubsection{Anesthesia} \label{sec:anesthesia}
Another major drug dosing problem in ICUs is the regulation and automation of sedation and analgesia, which is essential in maintaining physiological stability and lowering pains of patients. Whereas surgical patients typically require deep sedation over a short duration of time, sedation for ICU patients, especially when using mechanical ventilation, can be more challenging \cite{prasad2017reinforcement}. Critically ill patients who are supported by mechanical ventilation require adequate sedation for several days to guarantee safe treatment in the ICU \cite{haddad2012clinical}. A misdosing of sedation or under sedation is not acceptable since over sedation can cause hypotension, prolonged recovery time, delayed weaning from mechanical ventilation, and other related negative outcomes, whereas under sedation can cause symptoms such as anxiety, agitation and hyperoxia \cite{padmanabhan2015closed}.

The regulation of sedation in ICUs using RL methods has attracted attention of researcher for decades. As early as in 1994, Hu \emph{et al.} \cite{hu1994comparison} studied the problem of anesthesia control by applying some of the founding principles of RL (the MDP formulation and its planning solutions). More recently, RL-based control methods, using surrogate measures of anesthetic effect, e.g., the \emph{bispectral} (BIS) index, as the controlled variable, has enhanced individualized anesthetic management, resulting in the overall improvement of patient outcomes when compared with traditional controlled administration. Moore \emph{et al.} \cite{moore2004intelligent,sinzinger2005sedation} applied TD $Q(\lambda)$ in administration of intravenous propofol in ICU settings, using the well-studied Marsh-Schnider pharmacokinetic model to estimate the distribution of drug within the patient, and a pharmacodynamic model for estimating drug effect. The RL method adopted the error of BIS and estimation of the four compartmental propofol concentrations as the input state, different propofol dose as control actions, and the BIS error as the reward. The method demonstrated superior stability and responsiveness when compared to a well-tuned PID controller. The authors then modeled the drug disposition system as three states corresponding to the change of BIS, and applied basic Q-learning method to solving this problem \cite{moore2011reinforcement1,moore2011reinforcement2}. They also presented the first clinical \emph{in vivo}  trial for closed-loop control of anesthesia administration using RL on 15 human volunteers \cite{moore2014reinforcement,moore2010reinforcement}. It was demonstrated that patient specific control of anesthesia administration with improved control accuracy as compared to other studies in the literature could be achieved both in simulation and the clinical study.

Targeting at both muscle relaxation (paralysis) and \emph{Mean Arterial Pressure} (MAP), Sadati \emph{et al.} \cite{sadati2006multivariable} proposed an RL-based fuzzy controllers architecture in automation of the clinical anesthesia. A multivariable anesthetic mathematical model was presented to achieve an anesthetic state using two anesthetic drugs of \emph{Atracurium} and \emph{Isoflurane}. The highlight was that the physician's clinical experience could be incorporated into the design and implementation of the architecture, to realize reasonable initial dosage and keep drug inputs in safe values. Padmanabhan \emph{et al.} \cite{padmanabhan2015closed} used a closed-loop anesthesia controller to regulate the BIS and MAP within a desired range. Specifically, a weighted combination of the error of the BIS and MAP signals is considered in the proposed RL algorithm. This reduces the computational complexity of the RL algorithm and consequently the controller processing time. Borera \emph{et al.} \cite{borera2011adaptive} proposed an \emph{Adaptive Neural Network Filter} (ANNF) to improve RL control of propofol hypnosis.

Lowery and Faisal \cite{lowery2013towards} used a continuous AC method to first learn a generic effective control strategy based on average patient data and then fine-tune itself to individual patients in a personalization stage. The results showed that the reinforcement learner could reduce the dose of administered anesthetic agent by 9.4\% as compared to a fixed controller, and keep the BIS error within a narrow, clinically acceptable range 93.9\% of the time. More recently, an IRL method has been proposed that used expert trajectories provided by anesthesiologists to train an RL agent for controlling the concentration of drugs during a global anesthesia \cite{humbertlearning}.

\subsubsection{Other Applications in Critical Care} \label{sec:ICU_other}
While the previous sections are devoted to two topic-specific applications of RL methods in critical care domains, there are many other more general medical problems that perhaps have received less attention by researchers. One such problem is regarding the medication dosing, particulary, heparin dosing, in ICUs. A recent study by Ghassemi \emph{et al.} \cite{ghassemi2014data} highlighted that the misdosing of medications in the ICU is both problematic and preventable, e.g., up to two-thirds of patients at the study institution received a non-optimal initial dose of heparin, due to the highly personal and complex factors that affect the dose-response relationship. To address this issue, Nemati \emph{et al.} \cite{nemati2016optimal} inferred hidden states of patients via discriminative hidden Markov model and applied neural FQI to learn optimal heparin dosages. Lin \emph{et al.} \cite{lin2018deep} applied DDPG in continuous state-action spaces to learn a better policy for heparin dosing from observational data in MIMIC and the Emory University clinical data. Wang \emph{et al.} \cite{wang2018supervised} combined supervised signals and reinforcement signals to learn recommendations for medication dosing involving a large number of diseases and medications in ICUs.

Another typical application of RL in ICUs is to develop a decision support tool for automating the process of airway and mechanical ventilation. The need for mechanical ventilation is required when patients in ICUs suffer from \emph{acute respiratory failure} (ARF) caused by various conditions such as cardiogenic pulmonary edema, sepsis or weakness after abdominal surgery \cite{jaber2017intensive}. The management of mechanical ventilation is particularly challenging in ICUs. One one hand, higher costs occur if unnecessary ventilation is still taking effect, while premature extubation can give rise to increased risk of morbidity and mortality. Optimal decision making regarding when to wean patients off of a ventilator thus becomes nontrivial since there is currently no consistent clinical opinion on the best protocol for weaning of ventilation \cite{de2017focus}. Prasad \emph{et al.} \cite{prasad2017reinforcement} applied off-policy RL algorithms, FQI-ERT and with feed forward neural networks, to determine the best weaning time of invasive mechanical ventilation, and the associated personalized sedation dosage. The policies learned showed promise in recommending weaning protocols with improved outcomes, in terms of minimizing rates of reintubation and regulating physiological stability. Targeting at the same problem as \cite{prasad2017reinforcement}, Jagannatha \emph{et al.} \cite{jagannathatowards} analyzed the properties and limitations of standard off-policy evaluation methods in RL and discussed possible extensions to them in order to improve their utility in clinical domains. More recently, Yu \emph{et al.} applied Bayesian inverse RL \cite{yu2019inverse}  and Supervised-actor-critic \cite{yu2020supervised} to learn a suitable ventilator weaning policy from real trajectories in retrospective ICU data. RL has been also used in the development of optimal policy for the ordering of lab tests in ICUs \cite{cheng2018optimal,chang2018dynamic}, and prevention and treatments for \emph{graft versus host disease} (GVHD) \cite{krakow2017tools,liu2017deep} using data set from the \emph{Center for International Bone Marrow Transplant Research (CIBMTR)} registry database.

\section{Automated Medical Diagnosis}\label{sec:diagnosis}
Medical diagnosis is a mapping process from a patient's information such as treatment history, current signs and symptoms to an  accurate clarification of a disease. Being a complex task, medical diagnosis often requires ample medical investigation on the clinical situations, causing significant cognitive burden for clinicians to assimilate valuable information from complex and diverse clinical reports. It has been reported that diagnostic error accounts for as high as 10\% of deaths and 17\% of adverse events in hospitals \cite{national2016improving}. The error-prone process in diagnosis and the necessity to assisting the clinicians for a better and more efficient decision making urgently call for a significant revolution of the diagnostic process, leading to the advent of automated diagnostic era that is fueled by advanced big data analysis and machine learning techniques \cite{rai2018review,fatima2017survey,chui2017disease}.

Normally formulated as a supervised classification problem, existing machining learning methods on clinical diagnosis heavily rely on a large number of annotated samples in order to infer and predict the possible diagnoses \cite{lipton2015learning,choi2016retain,goodwin2016medical}. Moreover, these methods have limits in terms of capturing the underlying dynamics and uncertainties in the diagnosing process and considering only a limited number of prediction labels \cite{ling2017diagnostic}. To overcome these issues, researchers are increasingly interested in formulating the diagnostic inferencing problem as a sequential decision making process and using RL to leverage a small amount of labeled data with appropriate evidence generated from relevant external resources \cite{ling2017diagnostic}. The existing research can be classified into two main categories, according to the type of clinical data input into the learning process: the structured medical data such as physiological signals, images, vital signs and lab tests, and the unstructured data of free narrative text such as laboratory reports, clinical notes and summaries.

\subsection{Structured Medical Data}
The most successful application of RL in diagnosis using structured data pertains to various processing and analysis tasks in medical image examination, such as feature extracting, image segmentation, and object detection/localization/tracing \cite{bernstein2018reinforcement,taylor2004reinforcement}. Sahba \emph{et al.} \cite{sahba2006reinforcement,sahba2007application,sahba2008application,sahba2016object} applied basic Q-learning to the segmentation of the prostate in transrectal \emph{ultrasound images} (UI). Liu and Jiang  \cite{liu2018deep} used a DRL method, \emph{Trust Region Policy Optimization} (TRPO), for joint surgical gesture segmentation and classification. Ghesu \emph{et al.} \cite{ghesu2016artificial} applied basic DQN to automatic landmark detection problems, and achieved more efficient, accurate and robust performance than state-of-the-art machine learning and deep learning approaches on 2D\emph{ Magnetic Resonance Images} (MRI), UI and 3D \emph{Computed Tomography} (CT) images. This approach was later extended to exploit multi-scale image representations for large 3D CT scans \cite{ghesu2017multi}, and consider incomplete data \cite{ghesu2018towards} or nonlinear multi-dimensional parametric space in MRI scans of the brain region \cite{etcheverry2018nonlinear}.

Alansary \emph{et al.} evaluated different kinds of DRL methods (\emph{DQN}, \emph{Double DQN} (DDQN), \emph{Duel DQN}, and \emph{Duel DDQN}) \cite{li2018deep} for anatomical landmark localization in 3D fetal UI \cite{alansary2018evaluating}, and automatic standard view plane detection \cite{alansary2018automatic}. Al and Yun \cite{al2018partial} applied AC based direct PS method for aortic valve landmarks localization and left atrial appendage seed localization in 3D CT images. Several researchers also applied DQN methods in 3D medical image registration problems \cite{liao2017artificial,ma2017multimodal,krebs2017robust}, active breast lesion detection from dynamic contrast-enhanced MRI \cite{maicas2017deep}, and robust vessel centerline tracing problems in multi-modality 3D medical volumes \cite{zhang2018deep}.

Netto \emph{et al.} \cite{netto2008application} presented an overview of work applying RL in medical image applications, providing a detailed illustration of particular use of RL for lung nodules classification. The problem of classification is modeled as a sequential decision making problem, in which each state is defined as the combination of five 3D geometric measurements, the actions are random transitions between states, and the final goal is to discover the shortest path from the pattern presented to a known target of a malignant or a benign pattern. Preliminary results demonstrated that the Q-learning method can effectively classify lung nodules from benign and malignant directly based on lung lesions CT images.

Fakih and Das \cite{fakih2006lead} developed a novel RL-based approach, which is capable of suggesting proper diagnostic tests that optimize a multi-objective performance criterion accounting for issues of costs, morbidity, mortality and time expense. To this end, some diagnostic decision rules are first extracted from current medical databases, and then the set of possible testing choices can be identified by comparing the state of patient with the attributes in the decision rules. The testing choices and the combined overall performance criterion then serve as inputs to the core RL module and the VI algorithm is applied for obtaining optimized diagnostic strategies. The approach was evaluated on a sample diagnosing problem of \emph{solitary pulmonary nodule} (SPN) and results verified its success in improving testing strategies in diagnosis, compared with several other fixed testing strategies.

\subsection{Unstructured Medical Data}
Unlike the formally structured data that are directly machine understandable, large proportions of clinical information are stored in a format of unstructured free text that contains a relatively more complete picture of associated clinical events \cite{jiang2017artificial}. Given their expressive and explanatory power, there is great potential for clinical notes and narratives to play a vital role in assisting diagnosis inference in an underlying clinical scenario. Moreover, limitations such as knowledge incompleteness, sparsity and fixed schema in structured knowledge have motivated researchers to use various kinds of unstructured external resources such as online websites for related medical diagnosing tasks \cite{ling2017diagnostic}.

Motivated by the \emph{Text REtrieval Conference}-\emph{Clinical Decision Support} (TREC-CDS) track dataset \cite{roberts2016overview},  diagnosis inferencing from unstructured clinical text has gained much attention among AI researchers recently. Utilizing particular natural language processing techniques to extract useful information from clinical text, RL has been used to optimize the diagnosis inference procedure in several studies. Ling \emph{et al.} \cite{ling2017diagnostic,ling2017learning} proposed a novel clinical diagnosis inferencing approach that applied DQN to incrementally learn about the most appropriate clinical concepts that best describe the correct diagnosis by using evidences gathered from relevant external resources (from Wikipedia and MayoClinic). Experiments on the TREC-CDS datasets demonstrated the effectiveness of the proposed approach over several non RL-based systems.

Exploiting real datasets from the \emph{Breast Cancer Surveillance Consortium} (BCSC) \cite{ballard1997breast}, Chu \emph{et al.} \cite{chu2016adaptive} presented an adaptive online learning framework for supporting clinical breast cancer diagnosis. The framework integrates both supervised learning models for breast cancer risk assessment and RL models for decision-making of clinical measurements. The framework can quickly update relevant model parameters based on current diagnosis information during the training process. Additionally, it can build flexible fitted models by integrating different model structures and plugging in the corresponding parameters during the prediction process. The authors demonstrated that the RL models could achieve accurate breast cancer risk assessment from sequential data and incremental features.

In order to facilitate self-diagnosis while maintaining reasonable accuracy, the concept of \emph{symptom checking} (SC) has been proposed recently. SC first inquires a patient with a series of questions about their symptoms, and then attempts to diagnose some potential diseases \cite{goodwin2016medical}. Tang \emph{et al.} \cite{tang2016inquire} formulated inquiry and diagnosis policies as an MDP, and adopted DQN to learn to inquire and diagnose based on limited patient data. Kao \emph{et al.} \cite{kao2018context} applied context-aware HRL scheme to improve accuracy of SC over traditional systems making a limited number of inquiries. Empirical studies on a simulated dataset showed that the proposed model drastically improved disease prediction accuracy by a significant margin. The SC system was successfully employed in the DeepQ Tricorder which won the second prize in the Qualcomm Tricorder XPRIZE competition in year 2017 \cite{chang2017artificial,chang2017deepq}.

A dialogue system was proposed in \cite{wei2018task} for automatic diagnosis, in which the medical dataset was built from a pediatric department in a Chinese online healthcare community. The dataset consists of self-reports from patients and conversational data between patients and doctors. A DQN approach was then used to train the dialogue policy. Experiment results showed that the RL-based dialogue system was able to collect symptoms from patients via conversation and improve the accuracy for automatic diagnosis. In order to increase the efficiency of the dialogue systems, Tang \emph{et al.} \cite{tang2018improving} applied DQN framework to train an efficient dialogue agent to sketch disease-specific lexical probability distribution, and thus to converse in a way that maximizes the diagnosis accuracy and minimizes the number of conversation turns. The dialogue system was evaluated on the mild cognitive impairment diagnosis from a real clinical trial, and results showed that the RL-driven framework could significantly outperform state-of-the-art supervised learning approaches using only a few turns of conversation.

\section{Other Healthcare Domains} \label{sec:other_domains}
Besides the above applications of RL in DTR design and automated medical diagnosis, there are many other case applications in broader healthcare domains that focus on problems specifically in health resource scheduling and allocation, optimal process control, drug discovery and development, as well as health management.

(1) \emph{Health Resource Scheduling and Allocation}. The healthcare system is a typical service-oriented system where customers (e.g., patients) are provided with service using limited resources, e.g. the time slots, nursing resources or diagnostic devices \cite{schuetz2012approximate}. \emph{Business process management} (BPM) plays a key role in such systems as the objective of the service provider is to maximize profit overtime, considering various customer classes and service types with dynamics or uncertainties such as cancellations or no-shows of patients \cite{huang2011reinforcement,zeng2010clinic}. Since the optimal resource allocation problem in BPM can be seen as a sequential decision making problem, RL is then naturally suitable for offering reasonable solutions. Huang \emph{et al.} \cite{huang2011reinforcement} formulated the allocation optimization problems in BPM as an MDP and used basic Q-learning algorithm to derive an optimal solution. The RL-based approach was then applied to address the problem of optimizing resource allocation in radiology CT-scan examination process. A heuristic simulation-based approximate DP approach was proposed in \cite{schuetz2012approximate}, which considered both stochastic service times and uncertain future arrival of clients. The experimental investigation using data from the radiological department of a hospital indicated an increases of 6.9\% in the average profit of the hospital and 9\% in the number of examinations. Gomes \cite{gomes2017reinforcement} applied a DRL method, \emph{Asynchronous Advantage Actor Critic} (A3C) \cite{mnih2016asynchronous}, to schedule appointments in a set of increasingly challenging environments in primary care systems.

(2)\emph{ Optimal Process Control}. RL has also been  widely applied in deriving an optimal control policy in a variety of healthcare situations, ranging from surgical robot operation \cite{li2017function,thananjeyan2017multilateral,nguyen2019new,chen2016towards,baek2018path}, \emph{functional electrical stimulation} (FES) \cite{li2018inverse,jagodnik2017training}, and adaptive rate control for medical video streaming \cite{istepanian2009medical,alinejad2012cross}. Li and Burdick \cite{li2017function} applied RL to learn a control policy for a surgical robot such that the robot can conduct some basic clinical operations automatically. A function approximation based IRL method was used to derive an optimal policy from experts' demonstrations in high dimensional sensory state space. The method was applied to the evaluation of surgical robot operators in three clinical tasks of knot tying, needling passing and suturing. Thananjeyan \emph{et al.} \cite{thananjeyan2017multilateral} and Nguyen \emph{et al.} \cite{nguyen2019new}  applied DRL algorithm, TRPO, in learning tensioning policies effectively for  surgical gauze cutting. Chen \emph{et al.}  \cite{chen2016towards} combined programming by demonstration and RL for motion control of flexible manipulators in minimally invasive surgical performance, while Baek \emph{et al.} \cite{baek2018path} proposed the use of RL to perform resection automation of cholecystectomy by planning a path that avoids collisions in a laparoscopic surgical robot system.

FES employs neuroprosthesis controllers to apply electrical current to the nerves and muscles of individuals with spinal cord injuries for rehabilitative movement \cite{ragnarsson2008functional}. RL has been used to calculate stimulation patterns to efficiently adapt the control strategy to a wide range of time varying situations in patients' preferences and reaching dynamics. AC-based control strategies \cite{thomas2008creating,thomas2009application,jagodnik2017training} were proposed to evaluate target-oriented task performed using a planar musculoskeletal human arm in FES. To solve the reward learning problem in large state spaces, an IRL approach was proposed in \cite{li2018inverse} to evaluate the effect of rehabilitative stimulations on patients with spinal cord injuries based on the observed patient motions.

RL-based methods have also been widely applied in adaptive control in mobile health medical video communication systems. For example, Istepanian \emph{et al.} \cite{istepanian2009medical} proposed a new rate control algorithm based on Q-learning that satisfies medical quality of service requirements in bandwidth demanding situations of ultrasound video streaming. Alinejad \cite{alinejad2012cross} applied Q-learning for cross-layer optimization in real-time medical video streaming.

(3) \emph{Drug Discovery and Development}. Drug discovery and development is a time-consuming and costly process that usually lasts for 10-17 years, but with as low as around 10\% overall probability of success \cite{kola2004can}. To search an effective molecule that meets the multiple criteria such as bioactivity and synthetic accessibility in a prohibitively huge synthetically feasible molecule space is extremely difficult. By using computational methods to virtually design and test molecules, \emph{de novo} design offers ways to facilitate cycle of drug development \cite{schneider2013novo}. It is until recent years that RL methods have been applied in various aspects of \emph{de novo} design for drug discovery and development. Olivecrona  \cite{olivecrona2017molecular} used RL to fine tune the recurrent neural network in order to generate molecules with certain desirable properties through augmented episodic likelihood. Serrano \emph{et al.} \cite{serrano2018accelerating} applied DQN to solve the proteinligand docking prediction problem, while Neil \emph{et al.} \cite{neil2018exploring} investigated the PPO method in molecular generation. More recently, Popova \emph{et al.} \cite{popova2018deep} applied DRL methods to generate novel targeted chemical libraries with desired properties.

(4) \emph{Health Management}. As a typical application domain, RL has also been used in adaptive interventions to support health management such as promoting physical activities for diabetic patients  \cite{yom2017encouraging,hochberg2016reinforcement}, or weight management for obesity patients \cite{baniya2017adaptive,forman2018can}. In these applications, throughout continuous monitoring and communication of mobile health, personalized intervention policies can be derived to input the monitored measures and output when, how and which plan to deliver. A notable work was by Yom \emph{et al.} \cite{yom2017encouraging}, who applied RL to optimize messages sent to the users, in order to improve their compliance with the activity plan. A study of 27 sedentary diabetes type 2 patients showed that participants who received messages generated by the RL algorithm increased the amount of activity and pace of walking, while the patients using static policy did not. Patients assigned to the RL algorithm group experienced a superior reduction in blood glucose levels compared to the static control policies, and longer participation caused greater reductions in blood glucose levels.

\section{Challenges and Open Issues}\label{sec:challenges}
The content above has summarized the early endeavors and continuous progress of applying RL in healthcare over the past decades. Focus has been given to the vast variety of application domains in healthcare. While notable success has been obtained, the majority of these studies simply applied existing naive RL approaches in solving healthcare problems in a relatively simplified setting, thus exhibiting some common shortcomings and practical limitations. This section discusses several challenges and open issues that have not been properly addressed by the current research, from perspectives of how to deal with the basic components in RL (i.e., formulation of states, actions and rewards, learning with world models, and evaluation of policies), and fundamental theoretical issues in traditional RL research (i.e., the exploration-exploitation tradeoff and credit assignment problem).

\subsection{State/Action Engineering}
The first step in applying RL to a healthcare problem is determining how to collect and pre-process proper medical data, and summarize such data into some manageable state representations in a way that sufficient information can be retained for the task at hand. Selecting the appropriate level of descriptive information contained in the states is extremely important. On one hand, it would be better to contain as detailed information as possible in the states, since this complete information can provide a greater distinction among patients. On the other hand, however, increasing the state space makes the model become more difficult to solve. It is thus essential that a good state representation include any compulsory factors or variables that causally affect both treatment decisions and the outcomes. Previous studies have showed that, to learn an effective policy through observational medical data, the states should be defined in a way that to the most approximates the behavior policy that has generated such data \cite{gottesman2018evaluating,raghu2018behaviour}.

However, data in medical domains often exhibit notable biases or noises that are presumably varying among different clinicians, devices, or even medical institutes, reflecting comparable inter-patient variability \cite{vincent2014reinforcement}. For some complex diseases, clinicians still face inconsistent guides in selecting exact data as the state in a given case \cite{vellido2018machine}. In addition, the notorious issue of missing or incomplete data can further exaggerate the problem of data collection and state representation in medical settings, where the data can be collected from patients who may fail to complete the whole trial, or the number of treatment stages or timing of initializing the next line of therapy is flexible. This missing or censoring data will tend to increase the variance of estimates of the value function and thus the policy in an RL setting. While the missing data problem can be generally solved using various imputation methods that sample several possible values from the estimated distribution to fill in missing values, the censoring data problem is far more challenging, calling for more sophisticated techniques for state representation and value estimation in such flexible settings \cite{goldberg2012q,soliman2014personalized}.

Most existing work defines the states over the processed medical data with raw physiological, pathological, and demographics information, either using simple discretization methods to enable storage of value function in tabular form, or using some kinds of function approximation models (e.g., linear models or deep neural models). While this kind of state representation is simple and easy to implement, the rich temporal dependence or causal information, which is the key feature of medical data, can be largely neglected \cite{jeter2019does}. To solve this issue, various probabilistic graphical models \cite{koller2009probabilistic} can be used to allow temporal modeling of time series medical data, such as \emph{dynamic Bayesian networks} (DBNs), in which nodes correspond to the random variables of interest, edges indicate the relationship between these random variables, and additional edges model the time dependency. These kinds of graphical models have the desirable property that allows for interpretation of interactions between state variables or between states and actions, which is not the case for other methods such as SVMs and neural networks.

Coupled with the state representation in RL is the formulation of actions. The majority of existing work has mainly focused on discretization of the action space into limited bins of actions. Although this formulation is quite reasonable in some medical settings, such as choices in between turning on ventilation or weaning off it, there are many other situations where actions are by themselves continuous/multidimensional variables. While the simplification of discretizing medicine dosage is necessary in the early proof-of-concept stage, realizing fully continuous dosing in the original action space is imperative in order to meet the commitments of precision medicine \cite{jameson2015precision}. There has been a significant achievement in the continuous control using AC methods and PS methods in the past years, particularly from the area of robotic control \cite{kober2009policy} and DRL \cite{li2018deep}. While this achievement can provide direct solutions to this problem, selecting the action over large/infinite space is still non-trivial, especially when dealing with any sample complexity guarantees (PAC). An effective method for efficient action selection in continuous and high dimensional action spaces, while at the same time maintaining low exploration complexity of PAC guarantees would extend the applicability of current methods to more sample-critical medical problems.

\subsection{Reward Formulation}
Among all the basic components, the reward may be at the core of an RL process. Since it encodes the goal information of a learning task, a proper formulation of reward functions plays the most crucial role in the success of RL. However, the majority of current RL applications in healthcare domains are still grounded on simple numerical reward functions that must be explicitly defined beforehand to indicate the goal of treatments by clinicians. It is true that in some medical settings, the outcomes of treatments can be naturally generated and explicitly represented in a numerical form, for example, the time elapsed, the vitals monitored, or the mortality reduced. In general, however, specifying such a reward function precisely is not only difficult but sometimes even misleading. For instance, in treatment of cancers \cite{zhao2009reinforcement}, the reward function was usually decomposed into several independent or contradictory components based on some prior domain knowledge, each of which was mapped into some integer numbers, e.g.,  -60 as a high penalty for patient death and +15 as a bonus for a cured patient. Several threshold and weighting parameters were needed to provide a way for trading-off efficacy and toxicity, which heavily rely on clinicians' personal experience that varies from one to another. This kind of somewhat arbitrary quantifications might have significant influence on the final learned therapeutic strategies and it is unclear how changing these numbers can affect the resulting strategies.

To conquer the above limitations, one alternative is to provide the learning agent with more qualitative evaluations for actions, turning the learning into a PRL problem \cite{furnkranz2011preference}. Unlike the standard RL approaches that are restricted to numerical and quantitative feedback, the agent's preferences instead can be represented by more general types of preference models such as ranking functions that sort states, actions, trajectories or even policies from most to least promising \cite{wirth2017survey}. Using such kind of ranking functions has a number of advantages as they are more natural and easier to acquire in many applications in clinical practice, particularly, when it is easier to require comparisons between several, possibly suboptimal actions or trajectories than to explicitly specify their performance. Moreover, considering that the medical decisions always involve two or more related or contradictory aspects during treatments such as benefits versus associated cost, efficacy versus toxicity, and efficiency versus risk, it is natural to shape the learning problem into a multi-objective optimization problem. MORL techniques \cite{liu2015multiobjective} can be applied to derive a policy that makes a trade-off between distinct objectives in order to achieve a Pareto optimal solution. Currently, there are only very limited studies in the literature that applied PRL \cite{furnkranz2012preference,cheng2011preference,laber2014set} and MORL \cite{lizotte2016multi,lizotte2010efficient,lizotte2012linear,cheng2018optimal} in medical settings, for optimal therapy design in treatment of cancer, schizophrenia or lab tests ordering in ICUs. However, all these studies still focus on very limited application scenarios where only static preferences or fixed objectives were considered. In a medical context, the reward function is usually not a fixed term but subject to changing with regard to a variety of factors such as the time, the varying clinical situations and the evolving physiopsychic conditions of the patients. Applying PRL and MORL related principles to broader domains and considering the dynamic and evolving process of patients' preferences and treatment objectives is still a challenging issue that needs to be further explored.

A more challenging issue is regarding the inference of reward functions directly from observed behaviors or clinical data. While it is straightforward to formulate a reward function, either quantitatively or qualitatively, and then compute the optimal policy using this function, it is sometimes preferable to directly estimate the reward function of experts from a set of presumably optimal treatment trajectories in retrospective medical data. \emph{Imitation learning}, particularly, IRL \cite{ng2000algorithms,zhifei2012survey}, is one of the most feasible approaches to infer reward functions given observations
of optimal behaviour. However, applying IRL in clinical settings is not straightforward, due to the inherent complexity of clinical data and its associated uncertainties during learning. The variance during the policy learning and reward learning can amplify the bias in each learning process, potentially leading to divergent solutions that can be of little use in practical clinical applications \cite{herman2016inverse,asoh2013application}.

Last but not the least, while it is possible to define a short-term reward function at each decision step using prior human knowledge, it would be more reasonable to provide a long-term reward only at the end of a learning episode. This is especially the case in healthcare domains where the real evaluation outcomes (e.g., decease of patients, duration of treatment) can only be observed at the end of treatment. Learning with sparse rewards is a challenging issue that has attracted much attention in recent RL research. A number of effective approaches have been proposed, such as the \emph{hindsight experience replay} \cite{andrychowicz2017hindsight}, the \emph{unsupervised auxiliary learning} \cite{jaderberg2016reinforcement}, the \emph{imagination-augmented learning} \cite{racaniere2017imagination}, and the \emph{reward shaping} \cite{ng1999policy}. While there have been several studies that address the sparse reward problem in healthcare domains, most of these studies only focus on DTRs with a rather short horizon (typically three or four steps). Moreover, previous work has showed that entirely ignoring short-term rewards (e.g. maintaining hourly physiologic blood pressure for sepsis patients) could prevent from learning crucial relationships between certain states and actions \cite{jeter2019does}. How to tackle sparse reward learning with a long horizon in highly dynamic clinical environments is still a challenging issue in both theoretical and practical investigations of RL in healthcare.

\subsection{Policy Evaluation}
The process of estimating the value of a policy (i.e., target policy) with data collected by another policy (i.e., behavior policy) is called \emph{off-policy evaluation} problem \cite{sutton2018reinforcement}. This problem is critical in healthcare domains because it is usually infeasible to estimate policy value by running the policy directly on the target populations (i.e., patients) due to high cost of experiments, uncontrolled risks of treatments, or simply unethical/illegal humanistic concerns. Thus, it is needed to estimate how the learned policies might perform on retrospective data before testing them in real clinical environments. While there is a large volume of work in RL community that focuses on \emph{importance sampling} (IS) techniques and how to trade off between bias and variance in IS-based off-policy evaluation estimators (e.g., \cite{jiang2016doubly}),  simply adopting these estimators in healthcare settings might be unreliable due to issues of sparse rewards or large policy discrepancy between RL learners and physicians. Using sepsis management as a running example, Gottesman \emph{et al.}, \cite{gottesman2018evaluating} discussed in detail why evaluation of polices using retrospective health data is a fundamentally challenging issue. They argued that any inappropriate handling of state representation, variance of IS-based statistical estimators, and confounders in more ad-hoc measures would result in unreliable or even misleading estimates of the quality of a treatment policy. The estimation quality of the off-policy evaluation is critically dependent on how precisely the behaviour policy is estimated from the data, and whether the probabilities of actions under the approximated behaviour policy model represent the true probabilities \cite{raghu2018behaviour}. While the main reasons have been largely unveiled, there is still little work on effective policy evaluation methods in healthcare domains. One recent work is by Li \emph{et al.} \cite{li2018actor}, who provided an off-policy POMDP learning method to take into account uncertainty and history information in clinical applications. Trained on real ICU data, the proposed policy was capable of dictating near-optimal dosages in terms of vasopressor and intravenous fluid in a continuous action space for sepsis patients.

\subsection{Model Learning}

In the efficient techniques described in Section \ref{subsec:key_techniques}, model-based methods enable improved sample efficiency over model-free methods by learning a model of the transition and reward functions of the domain on-line and then planing a policy using this model \cite{hester2012learning}. It is surprising that there are quite limited model-based RL methods applied in healthcare in the current literature \cite{raghu2018model,komorowski2018artificial,utomo2018treatment}. While a number of model-based RL algorithms have been proposed and investigated in the RL community (e.g., R-max \cite{brafman2002r}, $E^3$ \cite{kearns2002near}), most of these algorithms assume that the agent operates in small domains with a discrete state space, which is contradictory to the healthcare domains usually involving multi-dimensional continuously valued states and actions. Learning and planning over such large scale continuous models would cause additional challenges for existing model-based methods \cite{hester2012learning}. A more difficult problem is to develop efficient exploration strategies in continuous action/state space \cite{li2012sample}. By deriving a finite representation of the system that both allows efficient planning and intelligent exploration, it is potential to solve the challenging model learning tasks in healthcare systems more efficiently than contemporary RL algorithms.

\subsection{Exploration Strategies}
Exploration plays a core role in RL, and a large amount of effort has been devoted to this issue in the RL community. A wealth of exploration strategies have been proposed in the past decades. Surprisingly, the majority of existing RL applications in healthcare domains simply adopt simple heuristic-based exploration strategies (i.e., $\varepsilon$-greedy strategy). While this kind of handling exploration dilemmas has made notable success, it becomes infeasible in dealing with more complicated dynamics and larger state/action spaces in medical settings, causing either a large sample complexity or an asymptotic performance far from the optimum. Particularly, in cases of an environment where only a rather small percentage of the state space is reachable, naive exploration from the entire space would be quite inefficient. This problem is getting more challenging in continuous state/action space, for instance, in the setting of HIV treatment \cite{ernst2006clinical}, where the basin of attraction of the healthy state is rather small compared to that of the unhealthy state. It has been shown that traditional exploration methods are unable to obtain obvious performance improvement and generate any meaningful treatment strategy even after a long period of search in the whole space \cite{pazis2013pac,kawaguchi2016bounded}. Therefore, there is a justifiable need for strategies that can identify dynamics during learning or utilize a performance measure to explore smartly in high dimensional spaces. In recent years, several more advanced exploration strategies have been proposed, such as PAC guaranteed exploration methods targeting at continuous spaces \cite{pazis2013pac,kawaguchi2016bounded}, concurrent exploration mechanisms \cite{pazis2016efficient,dimakopoulou2018coordinated,guo2015concurrent} and exploration in deep RL \cite{fu2017ex2,tang2017exploration,stadie2015incentivizing}. It is thus imperative to incorporate such exploration strategies in more challenging medical settings, not only to decrease the sample complexity significantly, but more importantly to seek out new treatment strategies that have not been discovered before.

Another aspect of applying exploration strategies in healthcare domains is the consideration of true cost of exploration. Within the vanilla RL framework, whenever an agent explores an inappropriate action, the consequent penalty acts as a negative reinforcement in order to discourage the wrong action. Although this procedure is appropriate for most situations, it may be problematic in some environments where the consequences of wrong actions are not limited to bad performance, but can result in unrecoverable effects. This is obviously true when dealing with patients in healthcare domains: although we can reset a robot when it has fallen down, we cannot bring back to life when a patience has been given a fatal medical treatment. Consequently, methods for safe exploration are of great real world interest in medical settings, in order to preclude unwanted, unsafe actions \cite{moldovan2012safe,garcia2015comprehensive,mannucci2018safe}.

\subsection{Credit Assignment}
Another important aspect of RL is the credit assignment problem that decides when an action or which actions is responsible for the learning outcome after a sequence of decisions. This problem is critical as the evaluation of whether an action being ``good'' or ``bad'' usually cannot be decided upon right away, but until the final goal has been achieved by the agent. As each action at each step contributes more or less to the final performance of success or failure, it is thus necessary to give distinct credit to the actions along the whole path, giving rise to the difficult problem of \emph{temporal credit assignment} problem. A related problem is the \emph{structural credit assignment} problem, in which the problem is to distribute feedback over the multiple candidates (e.g., multiple concurrently learning agents, action choices, or structure representations of the agent's policy).

The \emph{temporal credit assignment} problem is more prominent in healthcare domains as the effect of treatments can be much varied or delayed. Traditional RL research tackles the credit assignment problem using simple heuristics such as eligibility traces that weigh the past actions according to how far the time has elapsed (i.e., the backward view), or discount factors that weigh the future events according to how far away they will happen (i.e., the forward view) \cite{sutton2018reinforcement}. These kinds of fixed and simplified heuristics are incapable of modelling more complex interaction modes in a medical situation.  As a running example of explaining changes in blood glucose of a person with type 1 diabetes mellitus  \cite{merck2016causal}, it is difficult to give credit to the two actions of doing exercise in the morning or taking insulin after lunch, both of which can potentially cause hypoglycemia in the afternoon. Since there are many factors to affect blood glucose and the effect can take place after many hours, e.g., moderate exercise can lead to heightened insulin sensitivity for up to 22 hours, simply assigning an eligibility trace that decays with time elapsed is thus unreasonable, misleading or even incorrect. How to model the time-varying causal relationships in healthcare and incorporate them into the learning process is therefore a challenging issue that requires more investigations. The abundant literature in causal explanation \cite{woodward2005making} and inference \cite{morgan2015counterfactuals} can be introduced to provide a more powerful causal reasoning tool to the learning algorithm. By producing hypothesized sequences of causal mechanisms that seek to explain or predict a set of real or counterfactual events which have been observed or manipulated \cite{dash2013sequences}, not only can the learning performance be potentially improved, but also more explainable learned strategies can be derived, which is ultimately important in healthcare domains.

\section{Future Perspectives}\label{sec:future_perspectives}
We have discussed a number of major challenges and open issues raised in the current applications of RL techniques in healthcare domains. Properly addressing these issues are of great importance in facilitating the adoption of any medical procedure or clinical strategy using RL. Looking into the future, there is an urgent need in bringing recent development in both theories and techniques of RL together with the emerging clinical requirements in practice so as to generate novel solutions that are more interpretable, robust, safe, practical and efficient. In this section, we briefly discuss some of the future perspectives that we envision the most critical towards realizing such ambitions. We mainly focus on three theoretical directions: the interpretability of learned strategies, the integration of human or prior domain knowledge, and the capability of learning from small data. Healthcare under ambient intelligence and real-life applications are advocated as two main practical directions for RL applications in the coming age of intelligent healthcare.

\subsection{Interpretable Strategy Learning}
Perhaps one of the most profound issues with modern machine learning methods, including RL, is the lack of clear interpretability \cite{lipton2018mythos}. Usually functioning as a black box expressed by, for instance, deep neural networks, models using RL methods receive a set of data as input and directly output a policy which is difficult to interpret. Although impressive success has been made in solving challenging problems such as learning to play Go and Atari games, the lack of interpretability renders the policies unable to reveal the real correlation between features in the data and specific actions, and to impose and verify certain desirable policy properties, such as worst-case guarantees or safety constraints, for further policy debugging and improvement \cite{bhupatiraju2018towards}. These limits therefore greatly hinder the successful adoption of RL policies for safety-critical applications such as in medical domains as clinicians are unlikely to try new treatments without rigorous validation for safety, correctness and robustness  \cite{bastani2018verifiable,lipton2017doctor}.

Recently, there has been growing interest in attempting to address the problem of interpretability in RL algorithms. There are a variety of ways to realize interpretability of learned policy, by either using small, closed-form formulas to compute index-based policies \cite{maes2012policy}, using program synthesis to learn higher-level symbolic interpretable representations of learned policies \cite{verma2018programmatically}, utilizing genetic programming for interpretable policies represented by compact algebraic equations  \cite{hein2018interpretable}, or using program verification techniques to verify certain properties of the programs which are represented as decision trees \cite{bastani2018verifiable}.  Also, there has been growing attention not only on developing interpretable representations, but also on generating explicit explanations for sequential decision making problems \cite{elizalde2009generating}. While several works specifically focused on interpretability of deep models in healthcare settings \cite{che2015distilling,wu2018beyond}, how to develop interpretable RL solutions in order to increase the robustness, safety and correctness of learned strategies in healthcare domains is still an unsolved issue that calls for further investigations.

\subsection{Integration of Prior Knowledge}
There is a wealth of prior knowledge in healthcare domains that can be used for learning performance improvement. The integration of such prior knowledge can be conducted in different manners, either through configuration or presentation of learning parameters, components or models \cite{gaweda2005incorporating,gaweda2005individualization}, knowledge transfer from different individual patients, subtypes/sub-populations or clinical domains \cite{killian2017robust}, or enabling human-in-the-loop interactive learning \cite{holzinger2016interactive}.

Gaweda \emph{et al.} \cite{gaweda2005incorporating,gaweda2005individualization} presented an approach to management of anemia that incorporates a critical prior knowledge about the doseresponse characteristic into the learning approach, that is, for all patients, it is known that the dose-response curve of HGB vs. EPO is monotonically non-increasing. Thus, if a patient's response is evaluated as insufficient for a particular dose at a particular state, then the physician knows that the optimal dose for that state is definitely higher than the administered one. Consequently, there is no need to explore the benefit of lower doses at further stages of treatment. To capture this feature, the authors introduced an additional mechanism to the original Q-learning algorithm so that the information about monotonically increasing character of the HGB vs. EPO curve can be incorporated in the update procedure. This modification has been shown to make the EPO dosing faster and more efficiently.

While transfer learning has been extensively studied in the agent learning community \cite{taylor2009transfer}, there is quite limited work on applying TRL techniques in healthcare settings. The learning performance in the target task can be potentially facilitated by using latent variable models, pre-trained model parameters from past tasks, or directly learning a mapping between past and target tasks, thus extending personalized care to groups of patients with similar diagnoses. Marivate \emph{et al.} \cite{marivate2014quantifying} highlighted the potential benefit of taking into account individual variability and data limitations when performing batch policy evaluation for new individuals in HIV treatment.  A recent approach on TRL using latent variable models was proposed by Killian \emph{et al.} \cite{killian2016transfer,killian2017robust}, who used a Gaussian Process latent variable model for HIV treatment by both inferring the transition dynamics within a task instance and also in the transfer between task instances.

Another way of integrating prior knowledge into an RL process can be making use of the human cognitive abilities or domain expertise to guide, shape, evaluate or validate the agent's learning process, making the traditional RL into a human-in-the-loop interactive RL problem \cite{abel2017agent}. Human knowledge-driven RL methods can be of great interest to problems in healthcare domains, where traditional learning algorithms would possibly fail due to issues such as insufficient training samples, complex and incomplete data or unexplainable learning process \cite{holzinger2016interactive}. Consequently, the integration of the humans (i.e., doctors) into the learning process, and the interaction of an expert's  knowledge with the automatic learning data would greatly enhance the knowledge discovery process \cite{topol2019high}. While there is some previous work from other domains, particularly in training of robots \cite{yu2018adaptively,griffith2013policy}, human-in-the-loop interactive RL is not yet well established in the healthcare domain. It remains open for future research to transfer the insights from existing studies into the healthcare domain to ensure successful applications of existing RL methods.

\subsection{Learning from Small Data}
There is no doubt that the most recent progresses of RL, particularly DRL, are highly dependent on the premise of large number of training samples. While this is quite reasonable conceptually, that is, we cannot learn new things that we have not tried sufficiently enough, there still exist many domains lacking sufficient available training samples, specifically, in some healthcare domains \cite{shu2018small}. For example, in diagnose settings, medical images are much more difficult to be annotated with certain lesions in high-quality without specific expertise compared to general images with simple categories. In addition, there are usually few historical data or cases for new diseases and rare illness, making it impossible to obtain sufficient training samples with accurate labels. In such circumstances, directly applying existing RL methods on limited data may result in overly optimistic, or in other extreme, pessimistic about treatments that are rarely performed in practice.

Broadly, there are two different ways of dealing with a small sample learning problem \cite{carden2017small}. The direct solution can be using data augmentation strategies such as deformations \cite{salamon2017deep} or GANs \cite{goodfellow2014generative} to increase samples and then employ conventional learning methods. The other type of solutions can be applying various model modification or domain adaptation methods such as knowledge distillation \cite{hinton2015distilling} or meta-learning \cite{lake2017building} to enable efficient learning that overcomes the problem of data scarcity. While still in its early stage, significant progress has been made in small sample learning research in recent years \cite{carden2017small}. How to build on these achievements and tackle the small data RL problems in healthcare domains thus calls for new methods of future investigations. One initial work is by Tseng \emph{et al.} \cite{tseng2017deep} who developed automated radiation adaptation protocols for NSCLC patients by using GAN to generate synthetic patient data and DQN to learn dose decisions with the synthesized data and the available real clinical data. Results showed that the learned dose strategies by DQN were capable of achieving similar results to those chosen by clinicians, yielding feasible and quite promising solutions for automatic treatment designs with limited data.

\subsection{Healthcare under Ambient Intelligence}
The recent development in sensor networks and wearable devices has facilitated the advent of new era of healthcare systems that are characterized by low-cost mobile sensing and pervasive monitoring within the home and outdoor environments \cite{zheng2014unobtrusive}. The \emph{Ambient Intelligence} (AmI) technology, which enables innovative human-machine interactions through unobtrusive and anticipatory communications, has the potential to enhance the healthcare domain dramatically by learning from user interaction, reasoning reasoning about users' goals and intensions, and planning activities and future interactions \cite{acampora2013survey}. By using various kinds of sensing devices, such as smart phones, GPS and body sensors monitoring motions and activities, it is now possible to remotely and continuously collect patients' health information such that proper treatment or intervention decisions can be made anytime and anywhere.

As an instance of online decision making in a possibly infinite horizon setting involving many stages of interventions, RL plays a key role in achieving the future vision of AmI in healthcare systems through continuous interaction with the environment and adaption to the user needs in a transparent and optimal manner. In fact, the high level of monitoring and sensing provides ample opportunity for RL methods that can fuse estimates of a given physiologic parameter from multiple sources to provide a single measurement, and derive optimal strategies using these data. Currently, there are several studies that have applied RL to achieve AmI in healthcare domains. For example, RL has been used to adapt the intervention strategies of smart phones in order to recommend regular physical activity to people who suffer from diabetes type 2 \cite{yom2017encouraging,hochberg2016reinforcement}, or who have experienced a cardiac event and been in cardiac rehab \cite{zhu2018robust,zhu2018group,lei2014actor}. It has been also been used for mobile health intervention for college students who drink heavily and smoke cigarettes \cite{murphy2016batch}.

Despite the successes, healthcare under AmI poses some unique challenges that preclude direct application of existing RL methodologies for DTRs. For example, it typically involves a large number of time points or infinite time horizon for each individual; the momentary signal may be weak and may not directly measure the outcome of interest; and estimation of optimal treatment strategies must be done online as data accumulate. How to tackle these issues is of great importance in the successful applications of RL methods in the advent of healthcare systems under AmI.

\subsection{Future \emph{in-vivo} Studies}

To date, the vast volume of research reporting the development of RL techniques in healthcare is built upon certain computational models that leverages mathematical representation of how a patient responds to given treatment policies, or upon retrospective   clinical data to directly derive appropriate treatment strategies. While this kind of \emph{in silico} study is essential as a tool for early stage exploration or direct derivation of adaptive treatment strategies by providing approximate or highly simplified models, future \emph{in vivo} studies of closed-loop RL approaches are urgently required to reliably assess the performance and personalization of the proposed approaches in real-life implementations. However, a number of major issues still remain related to, in particular, data collection and preprocessing in real clinical settings, and high inter-individual differences of the physiological responses, thus calling for careful consideration of safety, efficiency and robustness of RL methods in real-life healthcare applications.

First and foremost, safety is of paramount importance in medical settings, thus it is imperative to ensure that the actions during learning be safe enough when dealing with \emph{in vivo} subjects. In some healthcare domains, the consequences of wrong actions are not merely limited to bad performance, but may include long-term effects that cannot be compensated by more profitable exploitation later on. As one wrong action can result in unrecoverable effects, learning in healthcare domains poses a safety exploration dilemma \cite{mannucci2018safe,moldovan2012safe}. It is worth noting that there are substantial ongoing efforts in the computer science community to address precisely these problems, namely in developing risk-directed exploration algorithms that can efficiently learn with formal guarantees regarding the safety (or worst-case performance) of the system \cite{garcia2015comprehensive}. With this consideration, the agent's choice of actions is aided by an appropriate risk metric acting as an exploration bonus toward safer regions of the search space. How to draw on these achievements and develop safe exploration strategies is thus urgently required to implement RL methods in real-life healthcare applications.

Another issue is regarding the sample efficiency of RL methods in \emph{in vivo} studies  \cite{zhao2009reinforcement}. While it is possible for the RL algorithms to collect large numbers of samples in simulations, it is unrealistic for sample-critical domains where collecting samples would cause significant cost. This is obviously true when dealing with real patients who would possibly not survive the long-term repeated trail-and-error treatment. Luckily, the wide range of efficient techniques reviewed in Section \ref{subsec:key_techniques} can provide promising solutions to this problem. Specifically, the sample-level batch learning methods can be applied for more efficient use of past samples, while model-based methods enable better use of samples by building the model of the environment. Another appealing solution is using task-level transfer methods that can reuse the past treatment or patient information to facilitate learning in new cases, or directly transfer the learned policies in simulations to real environments. To enable efficient transfer, RL algorithms can be provided with initial knowledge that can direct the learning in its initial stage toward more profitable and safer regions of the state space, or with demonstrations and teacher advice from an external expert that can interrupt exploration and provide expert knowledge when the agent is confronted with unexpected situations.

The last issue in the real-life implementation of RL approaches is regarding the robustness of derived solutions. Despite inherently being suitable for optimizing outcomes in stochastic processes with uncertainty, existing RL methods are still facing difficulties in handling incomplete or noisy state variables in partially observable real healthcare environments, and in providing measures of confidence (e.g. standard errors, confidence sets, hypothesis tests). Uncertainty can also be caused by the MDP parameters themselves, which leads to significant increases in the difficulty of the problem, in terms of both computational complexity and data requirements. While there has been some recent work on robust MDP solutions which accounts for this issue \cite{wiesemann2013robust,xu2010distributionally}, a more general and sound theoretical and empirical evaluation is still lacking. Moreover, most current studies are built upon predefined functions to map states and actions into some integer numbers. It is unclear how changing these numbers would affect the resulting optimal solutions. Understanding the robustness of RL methods in uncertain healthcare settings is the subject of ongoing critical investigations by the statistics, computer science and healthcare communities.

\section{Conclusions}\label{sec:conclusion}
RL presents a mathematically solid and technically sound solution to optimal decision making in various healthcare tasks challenged with noisy, multi-dimensional and incomplete data, nonlinear and complex dynamics, and particularly, sequential decision precedures with delayed evaluation feedback. This paper aims to provide a state-of-the-art comprehensive survey of RL applications to a variety of decision making problems in the area of healthcare. We have provided a structured summarization of the theoretical foundations and key techniques in the RL research from traditional machine learning perspective, and surveyed the broad-ranging applications of RL methods in solving problems affecting manifold areas of healthcare, from DTRs in chronic diseases and critical care, automated clinical diagnosis, to other healthcare domains such as clinical resource allocation and scheduling. The challenges and open issues in the current research have been discussed in detail from the perspectives of basic components constituting an RL process (i.e., states, actions, rewards, policies and models), and fundamental issues in RL research (i.e., the exploration-exploitation dilemma and credit assignment). It should be emphasized that, although each of these challenging issues has been investigated extensively in the RL community for a long time, achieving remarkably successful solutions, it might be problematic to directly apply these solutions in the healthcare settings due to the inherent complexity in processes of medical data processing and policy learning. In fact, the unique features embodied in the clinical or medical decision making process urgently call for development of more advanced RL methods that are really suitable for real-life healthcare problems. Apart from the enumerated challenges, we have also pointed out several perspectives that remain comparatively less addressed by the current literature. Interpretable learning, transfer learning as well as small-data learning are the three theoretical directions that require more effort in order to make substantial progress. Moreover, how to tailor the existing RL methods to deal with the pervasive data in the new era of AmI healthcare systems and take into consideration safety, robustness and efficiency caused by real-life applications are two main paradigms that need to be carefully handled in practice.

The application of RL in healthcare is at the intersection of computer science and medicine. Such cross-disciplinary research requires a concerted effort from machine learning researchers and clinicians who are directly involved in patient care and medical decision makings. While notable success has been obtained, RL has still received far less attention by researchers, either from computer science or from medicine, compared to other research paradigms in healthcare domains, such as traditional machine learning, deep learning, statistical learning and control-driven methods. Driven by both substantial progress in theories and techniques in the RL research, as well as practical demands from healthcare practitioners and managers, this situation is now changing rapidly and recent years have witnessed a surge of interest in the paradigm of applying RL in healthcare, which can be supported by the dramatic increase in the number of publications on this topic in the past few years. Serving as the first comprehensive survey of RL applications in healthcare, this paper aims at providing the research community with systematic understanding of foundations, broad palette of methods and techniques available, existing challenges, and new insights of this emerging paradigm. By this, we hope that more researchers from various disciplines can utilize their expertise in their own area and work collaboratively to generate more applicable solutions to optimal decision makings in healthcare.

\ifCLASSOPTIONcompsoc
  \section*{Acknowledgments}
  This work is supported by the Hongkong Scholar Program under Grant XJ2017028.
\else
  \section*{Acknowledgment}
  This work is supported by the Hongkong Scholar Program under Grant XJ2017028.\fi

\ifCLASSOPTIONcaptionsoff
  \newpage
\fi

\bibliographystyle{IEEEtran}
\bibliography{Arxiv_RL_Health}

\end{document}